\documentclass{article} 
\usepackage{main,times}

\usepackage{amsmath,amsfonts,bm}

\def\Figref#1{Figure~\ref{#1}}

\def\Secref#1{Section~\ref{#1}}

\def\eqref#1{equation~\ref{#1}}

\def\Tabref#1{Table~\ref{#1}}

\def\Appref#1{Appendix~\ref{#1}}

\def\1{\bm{1}}

\DeclareMathAlphabet{\mathsfit}{\encodingdefault}{\sfdefault}{m}{sl}
\SetMathAlphabet{\mathsfit}{bold}{\encodingdefault}{\sfdefault}{bx}{n}

\usepackage{url}
\usepackage{graphicx}
\usepackage{natbib}
\usepackage{caption}
\usepackage{subcaption}
\usepackage{algorithm}
\usepackage{algorithmic}
\usepackage{booktabs}
\usepackage{multirow}
\usepackage[colorlinks=true, backref=page, citecolor=teal]{hyperref}

\title{Understanding Decision-Making Mechanisms in Neural Routing Solvers}

\author{
  Fatemeh Askari\thanks{Equal contribution.} \quad
  Mazdak Teymourian\footnotemark[1] \quad
  Mohammad Izadi \quad
  Mahdieh Soleymani Baghshah \\[0.5em]
  Department of Computer Engineering \\
  Sharif University of Technology \\[0.3em]
  \texttt{\{fatemehaskarijirhandeh, mazdak.tey\}@gmail.com} \\
  \texttt{\{izadi, soleymani\}@sharif.edu}
}

\iclrfinalcopy 
\begin{document}

\maketitle
\makeatletter
\lhead{\small Preprint.}
\makeatother

\begin{abstract}
    Neural Combinatorial Optimization (NCO) has achieved strong empirical success, yet the internal mechanisms driving model decisions remain largely unexplored. In this paper, we investigate three representative autoregressive NCO models spanning two encoder-decoder configurations: AM and POMO (heavy-encoder, light-decoder), and LEHD (light-encoder, heavy-decoder). Through behavioral analyses, representation probing, and causal interventions, we examine how these models construct solutions and use internal representations during decoding. Our results suggest that AM and POMO predominantly follow a persistent geometric pattern throughout solution construction, whereas LEHD contains linearly accessible information about multiple future actions. Causal experiments further provide evidence for the role of future-node representations in LEHD's decision-making. We also observe that LEHD relies strongly on the current-node representation for immediate local decisions, while the start-node representation plays a broader navigational role over the subsequent route. Cross-instance alignment analyses additionally indicate that LEHD maps current-node representations into a relatively shared latent region, which may provide a stable reference for evaluating subsequent decisions. Across the Traveling Salesman Problem and the Capacitated Vehicle Routing Problem, these results reveal distinct decision-making patterns across these architecturally distinct solvers and provide a foundation for more interpretable analyses of NCO solvers. Code and additional visualizations are provided in the \href{https://github.com/NCO-Interpretability/NCO-Interpretability}{https://github.com/NCO-Interpretability/NCO-Interpretability}.
\end{abstract}
\section{Introduction}\label{sec:introduction}

The Traveling Salesman Problem (TSP) \citep{grotschel1991solution} and the Vehicle Routing Problem (VRP) \citep{dantzig1959truck} are among the most celebrated combinatorial problems in computer science, with critical applications spanning transportation \citep{pillac2013review}, logistics \citep{grigorios2022vehicle}, and drug discovery \citep{liu2017combinatorial}. Due to the NP-hard nature of these routing problems, exact algorithms are computationally prohibitive at scale. While classical heuristics like Concorde \citep{applegate2006traveling} and LKH3 \citep{helsgaun2017extension} produce high-quality solutions, they incur heavy computational overhead. Consequently, Neural Combinatorial Optimization (NCO) has emerged as a significantly faster alternative capable of achieving near-optimal or superior performance on tailored distributions.

\begin{figure*}[t!]
    \centering
    \includegraphics[width=\linewidth]{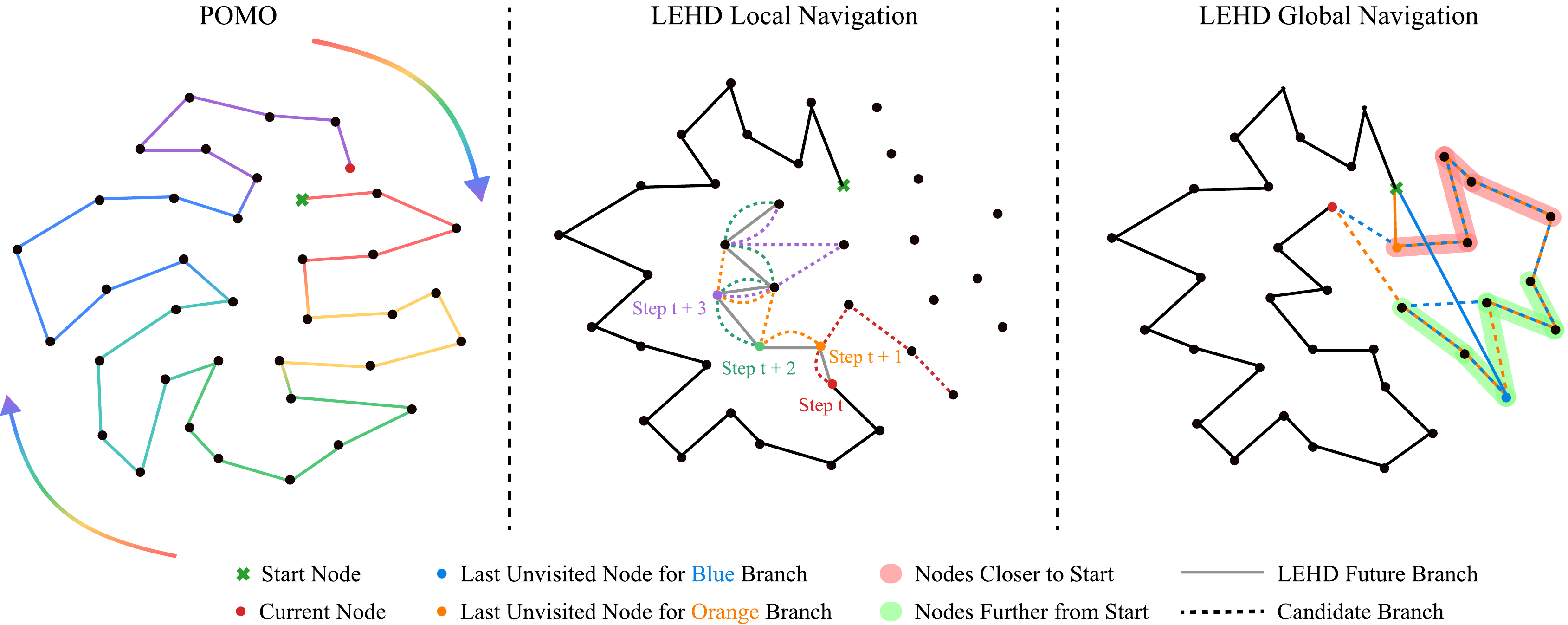}
    \caption{\textbf{Comparison of POMO and LEHD tour construction strategies.} 
    \textbf{Left:} POMO constructs the tour in a clockwise manner while oscillating between shallow and deep onion layers, with each oscillation highlighted in a distinct color. 
    \textbf{Middle:} LEHD local navigation across consecutive steps. Colored dashed lines represent candidate multi-step plans, which LEHD dynamically revises at each step. 
    \textbf{Right:} LEHD global navigation. LEHD evaluates candidate branches based on the relative angular position of the start node relative to the current node. It prioritizes the branch visiting nodes further from the start first (\textbf{orange branch}) over the branch visiting closer nodes first (\textbf{blue branch}), minimizing the final return cost to the start node.}
    \label{fig:pomo_lehd_abstract}
\end{figure*}

Among NCO approaches, Transformer-based architectures \citep{vaswani2017attention} represent a dominant paradigm. Trained via Reinforcement Learning (RL), AM \citep{kool2019attention} employs a heavy encoder and a lightweight decoder (HELD) to construct tours by sequentially appending unvisited nodes to the current partial path. For each instance, AM uses a learned policy to choose the starting node. POMO \citep{kwon2021pomo} also uses a HELD architecture and RL to train the model but it also leverages the rotational and structural symmetries inherent to cyclic routing problems and uses different starting positions for each instance, thus not requiring a policy to select a starting node. However, AM and POMO struggle to generalize to problem instances significantly larger than those seen during training. Addressing this bottleneck, LEHD \citep{luo2024neural} shifts to a Supervised Learning (SL) paradigm using an inverted architectural allocation, a light encoder (1 layer) paired with a heavy decoder (6 layers). At each step, LEHD receives the start node, current node, and unvisited set to predict the next step. Beyond autoregressive appending, \citet{luo2025learning} recently proposed L2C-Insert, an insertion-based heuristic that decouples generation into node selection and placement phases, using an encoder-decoder network to determine optimal insertion points within the partial tour.

Despite the rapid evolution of NCO architectures, the internal mechanisms driving their predictions remain largely black boxes. A few pioneering studies have begun peeling back these layers: \citet{zhang2025probing} used linear probes to demonstrate that Euclidean distances are linearly decodable from internal representations and identified specific embedding dimensions critical to LEHD's predictions, while \citet{narad2025mechanistic} applied Sparse Autoencoders (SAEs) to show that latent neurons capture geometric features such as boundary detection and spatial clustering. While insightful, these prior efforts offer limited mechanistic insights, focusing primarily on correlational analysis rather than establishing causality. Consequently, a fundamental question remains open: \textit{what explicit, macro-level solving strategies do these networks actually learn and execute?} To bridge this gap, we present a comprehensive interpretability framework, combining behavioral probing, representation analysis, activation patching, and causal analysis, to systematically dissect the strategies of leading NCO models. To the best of our knowledge, ours is the first work to employ causal interventions to explain the decision-making mechanisms of these solvers.

Using this framework, we analyze AM, POMO, and LEHD as representative construction paradigms, uncovering distinct spatial and mechanistic behaviors:
\begin{itemize}
    \item \textbf{AM and POMO:} Construct tours through a rigid spatial pattern, sweeping anti-clockwise for AM and clockwise for POMO from the initial node while systematically oscillating between shallow and deep convex (onion) layers (\Figref{fig:pomo_lehd_abstract}). As detailed in \Secref{sec:pomo}, this structural rigidity stems directly from training on uniform point distributions and shifts predictably when retrained on clustered instances.
    \item \textbf{LEHD:} Exhibits evidence of two complementary decision components which share similarities with Model Predictive Control (MPC) \citep{richalet1978model}. Locally, its final decoder layers encode explicit multi-step trajectories over an effective horizon, executing the immediate step before dynamically revising its trajectory (\Figref{fig:pomo_lehd_abstract}). Globally, LEHD utilizes the relative angular position of the start node to steer navigation toward distant unvisited clusters first, minimizing the eventual return cost. We also find that LEHD maps different current nodes to a relatively shared latent embedding space across instances and steps, suggesting that it plans in a canonical, current-centered latent reference frame.
\end{itemize}

Our main contributions are summarized as follows:
\begin{itemize}
    \item We conduct a diverse set of experiments to provide a mechanistic explanation of how distinct NCO architectures construct routing solutions.
    \item We characterize the geometric structure of the tours constructed by AM and POMO, demonstrating the rigidity of their learned strategy and how the training data distribution influences it.
    \item We reveal that LEHD coordinates a local look-ahead horizon with global spatial reference-frame navigation, explaining its strong generalization capabilities across problem scales.
\end{itemize}

Core findings generalize across both TSP and CVRP (details in \Appref{sec:cvrp-results}). Appendices~A–G cover related work, extended results, extra visualizations, and experimental setups.

\begin{figure*}[t]
    \centering
    \includegraphics[
        width=\textwidth
    ]{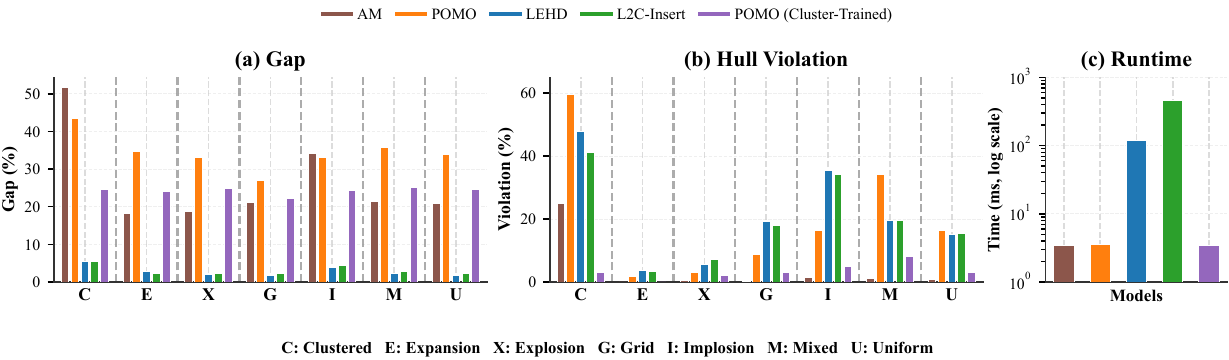}

    \caption{
Behavioral comparison of NCO solvers on TSP-500 under seven synthetic
distribution shifts. AM, POMO, LEHD, L2C-Insert, and cluster-trained POMO are
compared by optimality gap and convex-hull order violations across
Clustered, Expansion, Explosion, Grid, Implosion, Mixed, and Uniform
distributions, together with inference runtime.
The results highlight differences in solution quality, geometric consistency,
and computational efficiency under diverse instance geometries.
    }

    \label{fig:tsp500_behavioral_comparison}
\end{figure*}

\section{Behavioral Results and Model Comparison}
\label{sec:behavioural-model-comparison}

We compare AM, POMO, and LEHD as representative append-based solvers with different encoder--decoder allocations, and include L2C-Insert as an insertion-based reference. 
Throughout the paper, all models are evaluated using greedy, single-trajectory decoding without model-specific test-time enhancements, such as POMO's rotational augmentation or LEHD's RRC. 
This evaluation protocol allows us to study the intrinsic decision-making behavior of the learned policies without introducing additional inference-time modifications that may obscure the underlying model mechanisms.

All models are trained on Uniform TSP-100 and evaluated in
Figure~\ref{fig:tsp500_behavioral_comparison} on TSP-500 across seven
distributions, thereby testing both size and distribution generalization.
We report optimality gap, convex-hull violation rate, and inference runtime
on a logarithmic scale; a hull violation denotes failure to preserve the cyclic order of convex-hull vertices, a necessary property of an optimal Euclidean TSP tour.

L2C-Insert incurs substantially higher inference costs without consistently improving solution quality over LEHD.
LEHD achieves competitive solution quality across diverse distributions while maintaining substantially lower inference cost than the insertion-based baseline. 
Therefore, we retain L2C-Insert as an insertion-based reference for behavioral comparison, while focusing our mechanistic analyses on AM, POMO, and LEHD due to their shared append-based construction process, which enables a controlled comparison of their decision mechanisms. 
Additional details are provided in \Appref{sec:behavioral-extended}.

\begin{figure*}[t!]
    \centering
    \begin{subfigure}[t]{0.32\textwidth}
        \centering
        \includegraphics[width=\linewidth]{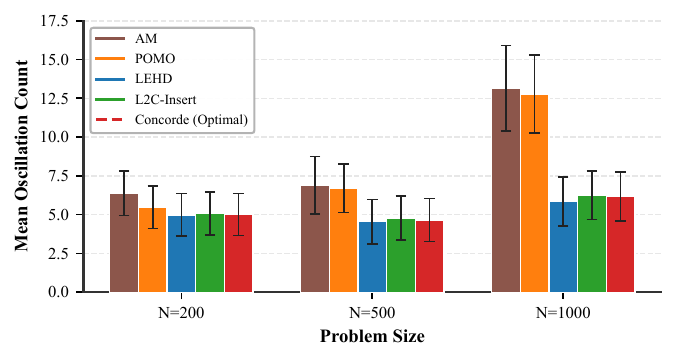}
        \caption{Mean onion layer oscillation}
        \label{fig:excursion-count-uniform}
    \end{subfigure}
    \begin{subfigure}[t]{0.32\textwidth}
        \centering
        \includegraphics[width=\linewidth]{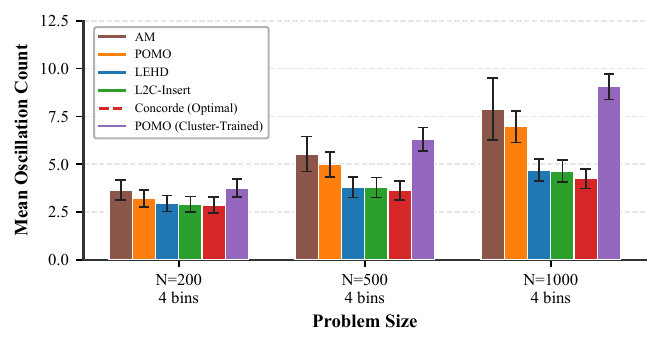}
        \caption{Mean intra-cluster oscillation}
        \label{fig:excursion-count-cluster}
    \end{subfigure}
    \begin{subfigure}[t]{0.32\textwidth}
        \centering
        \includegraphics[width=\linewidth]{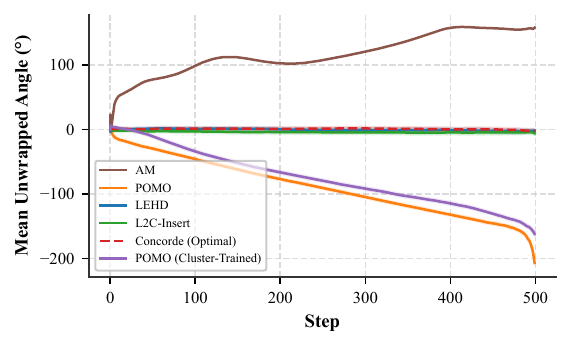}
        \caption{Angle comparison per step}
        \label{fig:angle-progression}
    \end{subfigure}
    \caption{\textbf{Tracking geometrical properties of generated tours.} 
    \textbf{(a)} Onion layers are segmented into $B$ bins to calculate the average number of oscillations between shallow and deeper layers, where higher values indicate a greater number of deep oscillations. 
    \textbf{(b)} Each cluster is partitioned into binwise onion layers to count high-intensity oscillations within individual clusters. 
    \textbf{(c)} The average start-to-current node angle relative to each tour's
    first edge is tracked throughout tour construction. A decreasing angle indicates a clockwise rotation of the start-to-current displacement vector.}
    \label{fig:polar_trajectory}
\end{figure*}
\section{Experimental Framework}\label{sec:experiments}
This section presents the experimental framework used to interpret NCO solvers. Following the problem formulations in \Appref{sec:preliminaries}, we first characterize their observable geometric behavior and then examine the representational and causal mechanisms underlying their decisions. The experimental settings are provided in \Appref{sec:experiment-settings} and the corresponding CVRP analyses are reported in \Appref{sec:cvrp-results}. These analyses are subsequently synthesized in \Secref{sec:pomo} and \Secref{sec:lehd} to develop an interpretation of the distinct decision-making strategies learned by AM, POMO, and LEHD.

\subsection{Geometric Trajectory Analysis}
\label{subsec:geometric-trajectory}

We characterize the generated solutions through two complementary geometric summaries: onion-depth progression and start-centered angular progression. The first measures how a solver moves across nested convex layers of the point set, while the second describes how selected nodes are distributed angularly with respect to the first generated edge.

\subsubsection{Onion-Depth Progression}
\label{subsubsec:onion-decomposition}

We use onion decomposition to characterize the layer-wise structure of TSP instances \citep{chazelle1985convex,compgeom2000}. It recursively removes the convex hull of the remaining points, partitioning them into nested layers $\rho_t$ (\Appref{subsec:onion-decomposition}). We then discretize \(\rho_t\in[0,1]\) into \(B\) equal-width bins
$b_t
=
1+
\min\left(
B-1,
\left\lfloor B\rho_t\right\rfloor
\right).$

Bin \(1\) is the outermost region and bin \(B\) the deepest. After removing consecutive duplicate bins, we count an oscillation when the trajectory leaves bin \(1\), reaches at least \(b_{\min}=\min\left(B,\left\lfloor \tau B\right\rfloor+1\right)\), and returns to bin \(1\), where \(\tau\in(0,1]\) controls the required depth. We report the average number of such oscillations across problem instances.

\subsubsection{Angular Progression}
\label{subsubsec:angular-progression}

We measure start-centered angular progression using the first generated edge as the reference axis,
\(a=x_{\pi_1}-x_{\pi_0}\), where \(\pi_0\) is the start node. At step \(t \in \{1,\ldots,N-1\}\), we compare it with
\(v_t=x_{\pi_t}-x_{\pi_0}\).
The signed angular position is computed as
$\theta_t
=
\operatorname{atan2}
\left(
a^{x}v_t^{y}-a^{y}v_t^{x},
a^\top v_t
\right).$

We unwrap \(\theta_t\) to remove artificial discontinuities at the \([-180^\circ,180^\circ]\) boundary, allowing the angular progression to vary continuously beyond this interval. We then compute the mean signed angle across instances at each decoding step, capturing systematic clockwise or counterclockwise progression around the start node.

For CVRP, angles are measured within each route relative to its first edge, excluding depot returns. Routes are aligned by normalized progress before aggregation.

\subsection{Future-Action Planning Probes}
\label{subsec:probing}

We use linear probes to examine whether neural combinatorial solvers encode information about their own future decisions. At decoding step \(t\), the frozen model has produced the partial solution
\(\pi_{\leq t}=[\pi_0,\pi_1,\ldots,\pi_t]\), and the current decoding state defines a feasible candidate set \(\mathcal{A}_t\). For TSP, this set consists of the unvisited nodes,
\(\mathcal{A}_t=V\setminus\{\pi_0,\ldots,\pi_t\}\). For routing problems with additional constraints, such as CVRP,
$\mathcal{A}_t$ contains all unvisited customers, regardless of
current capacity constraints. For a future horizon \(h\), the probing target is the action selected by the same frozen model at step \(t+h\) during its greedy rollout,
\(y_{t,h}=\pi_{t+h}\). The probe therefore does not predict the optimal solution; instead, it tests whether the model's current internal representation already contains information about its own future trajectory. For each feasible candidate \(i\in\mathcal{A}_t\), we extract a 128-dimensional candidate-specific representation \(r_{t,i}^{(\ell)}\) from internal layer \(\ell\). A separate linear probe is trained for each layer \(\ell\) and horizon \(h\), assigning one scalar score to every feasible candidate:
$s_{t,i}^{(\ell,h)}
=
W_{\ell,h}r_{t,i}^{(\ell)}
+
b_{\ell,h}.$
The probe is optimized using cross-entropy loss.

\begin{figure*}[t!]
    \centering
    \begin{subfigure}[t]{0.24\textwidth}
        \centering
        \includegraphics[width=\linewidth]{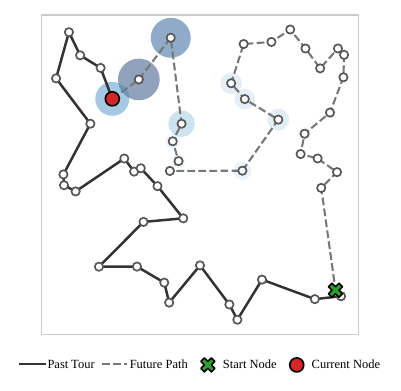}
        \caption{LEHD attention}
        \label{fig:lehd-attention-heatmap}
    \end{subfigure}%
    \hfill
    \begin{subfigure}[t]{0.43\textwidth}
        \centering
        \includegraphics[width=\linewidth]{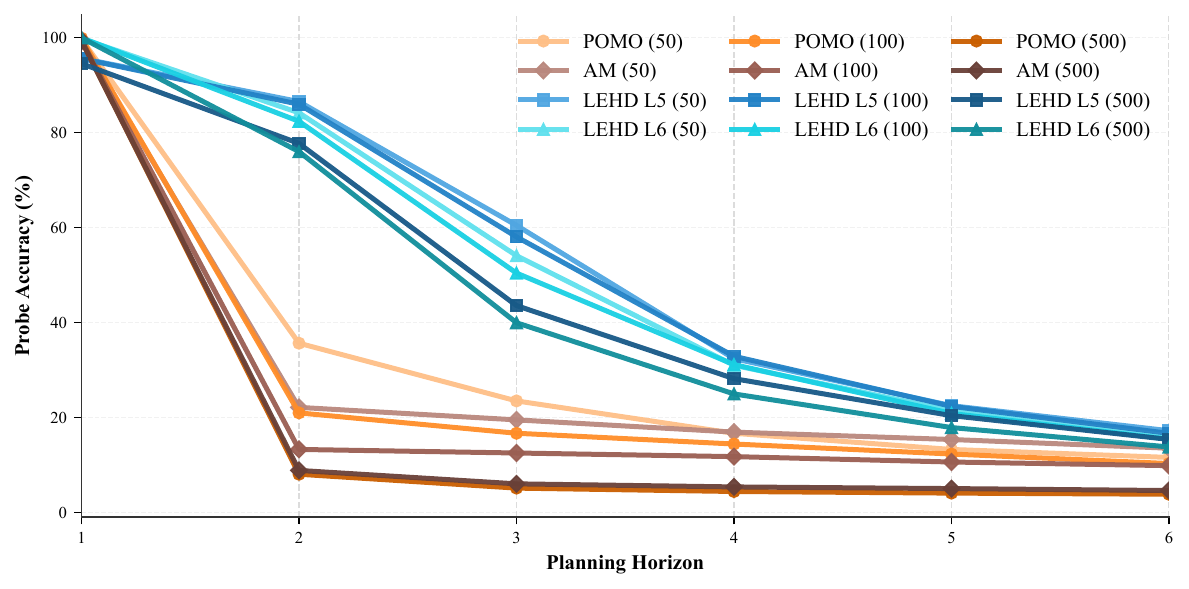}
        \caption{Probe accuracy per planning horizon}
        \label{fig:pomo_lehd_horizon_probe}
    \end{subfigure}%
    \hfill
    \begin{subfigure}[t]{0.32\textwidth}
        \centering
        \includegraphics[width=\linewidth]{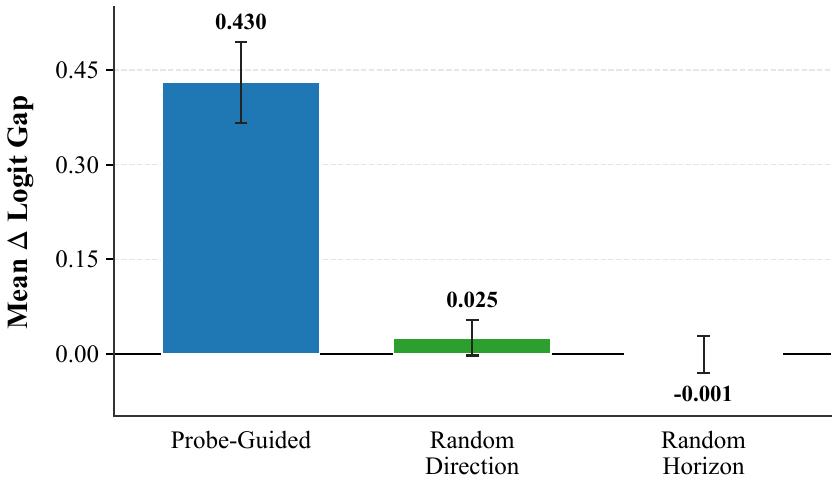}
        \caption{Probe-guided steering}
        \label{fig:lehd-probe-steer}
    \end{subfigure}
    \caption{\textbf{(a)} LEHD attention heatmap from the current node to the unvisited nodes.
    \textbf{(b)} Probe accuracy across planning horizons for the decoder layers of LEHD, AM, and POMO.
    \textbf{(c)} Probe-guided counterfactual steering for LEHD on TSP-100 at $\alpha=5$, compared against random-direction
    and random-horizon baselines. Bars show the mean change
    in logit gap between the clean top-two actions
    }
\end{figure*}

This formulation treats future-action prediction as a candidate-ranking problem. If a linear probe can recover the model's future action \(\pi_{t+h}\) from the representation at step \(t\), then that representation contains linearly accessible information about the model's subsequent decisions. We apply the same probing protocol to the  POMO, AM, and LEHD (\Figref{fig:pomo_lehd_horizon_probe}), while the model-specific representation choices are described in their corresponding analysis sections.

\subsection{Probe-Guided Counterfactual Steering}
\label{subsec:counterfactual-steering}

Inspired by prior work combining planning probes with causal
interventions and activation steering
\citep{bush2025interpreting,li2023inference,panickssery2023steering,li2022emergent,nanda2309emergent},
we investigate whether amplifying representations associated
with an alternative future route can shift the model's
immediate action preference.

At decoding step $t$, let $a$ and $a'$ denote the highest-
and second-highest-probability feasible actions, respectively.
We construct a counterfactual continuation by forcing
$\pi'_{t+1}=a'$ and then following the frozen model's greedy
policy. We ask whether intervening on representations of
nodes appearing at future steps along this counterfactual
continuation can causally shift the model's immediate action
preference toward $a'$.

For future horizons $h\in\mathcal{H}$ and decoder layers
$\ell\in\mathcal{L}$, we steer the representation of the
counterfactual future node $y'_{t,h}=\pi'_{t+h}$ along its
probe direction:
$\widetilde{r}_{t,y'_{t,h}}^{(\ell)}
=r_{t,y'_{t,h}}^{(\ell)}
+\alpha W_{\ell,h}^{\top}/\|W_{\ell,h}\|_2^2$,
where $\alpha>0$ controls intervention strength.
We jointly steer horizons $\mathcal{H}=\{2,3,4,5\}$,
without directly modifying the immediate candidates
$a$ or $a'$.

We then recompute the action logits without forcing $a'$
and measure the change in its preference over $a$:
$\Delta G_t=
[z_{\mathrm{steered}}(a')-z_{\mathrm{steered}}(a)]
-[z_{\mathrm{clean}}(a')-z_{\mathrm{clean}}(a)]$.
Controls use norm-matched random directions or an equal
number of disjoint random future horizons.

\subsection{Node-Role Representation Alignment}
\label{subsec:node-role-alignment}
We measure the cross-instance cosine similarity of node representations at each decoding step. For each layer, decoding step, and node role, we compare the corresponding representations across problem instances. 
This quantifies how consistently each functional role is represented across instances throughout decoding.

\subsection{Current and Start Node Contribution}
\label{subsec:activation-patching}

We use activation patching (\Appref{subsec:mechanistic_interpretability}) to examine the roles of two state-defining node
representations during decoding: the current node and the start node. The
current node represents the local position from which the next decision is
made, while the start node is a fixed element of the generated tour that
remains available to the decoder throughout the rollout. In both interventions
below, we modify the encoder-produced node embeddings before
they are passed to the decoder.

\subsubsection{Node Attribution}
\label{subsubsec:mean-ablation}

We perform stepwise mean ablations of the start and current node representations. At each decoding step, we compute a mean embedding over all clean encoded node representations collected from a reference set of instances at that step, and temporarily replace either the start-node or current-node representation with this mean. This removes instance-specific information while keeping the remaining decoding state unchanged. We then measure the drop in the clean next-node probability to assess the influence of each representation on the immediate decision.

\subsubsection{Targeted Intervention}\label{subsubsec:activation-transfer}
In the start-node patching experiment, we replace the encoded representation of the original start node with the representation of an already visited donor node, i.e., \(H'_{\pi_0}=H_d\), where \(\pi_0\) is the original start node and \(d\) is the donor node. This intervention changes the start-node representation while leaving the decoding state otherwise unchanged. We select donor nodes using three geometric conditions: maximum angle without distance constraints, maximum angle under similar distance, and similar angle with different distance. These conditions allow us to separate the effects of angular direction and distance in the start-node representation.

\begin{figure*}[t!]
    \centering
    \begin{subfigure}[t]{0.28\textwidth}
        \centering
        \includegraphics[width=\linewidth]{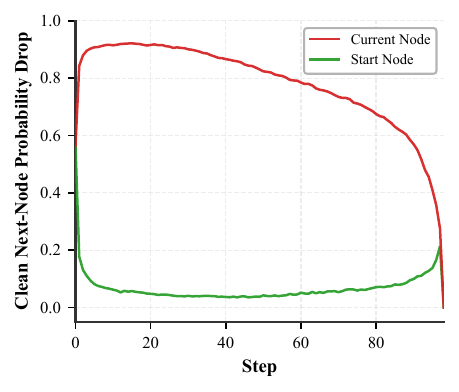}
        \caption{LEHD mean ablation}
        \label{fig:mean_ablation}
    \end{subfigure}
    \begin{subfigure}[t]{0.45\textwidth}
        \centering
        \includegraphics[width=\linewidth]{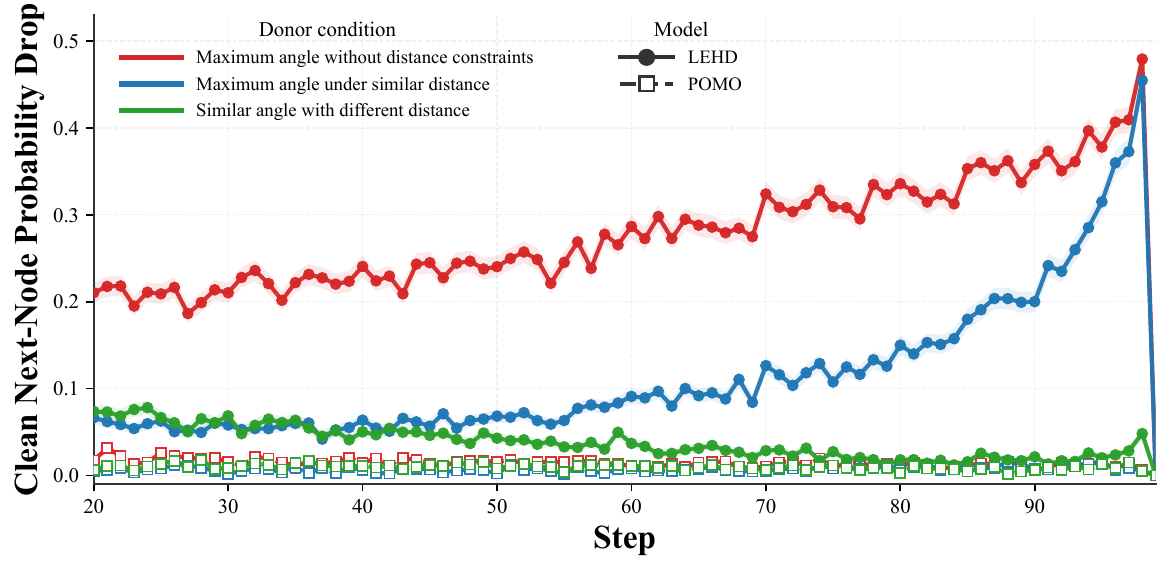}
        \caption{POMO vs. LEHD start-node patching}
        \label{fig:start_patching_plot}
    \end{subfigure}
    \begin{subfigure}[t]{0.23\textwidth}
        \centering
        \includegraphics[width=\linewidth]{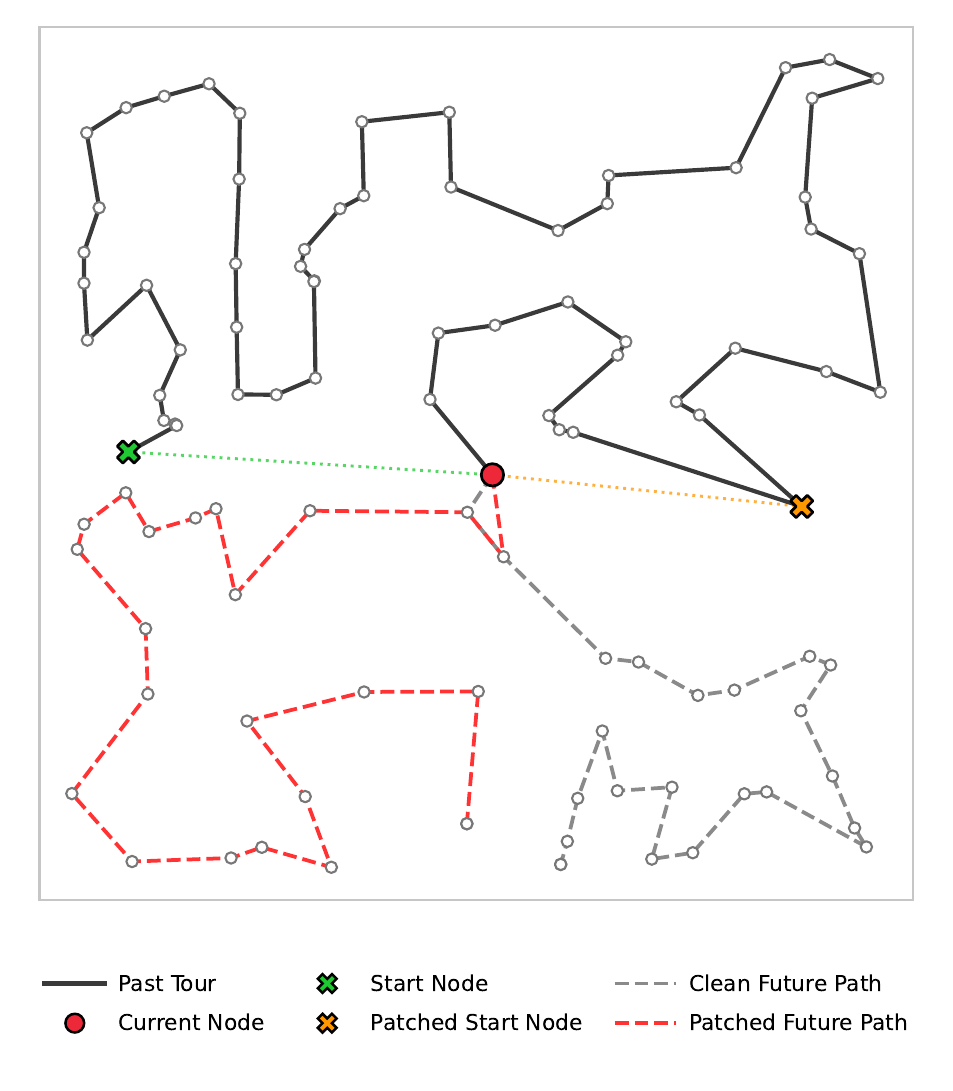}
        \caption{LEHD trajectory}
        \label{fig:trajectory_instance}
    \end{subfigure}
    \caption{
    Causal interventions on node representations.
    \textbf{(a)} Drop in the clean next-node probability after mean ablation of the current and start nodes.
    \textbf{(b)} Effect of start-node activation patching under three geometric donor conditions for LEHD and POMO.
    \textbf{(c)} Example clean and patched LEHD trajectories showing route changes over 22 decoding steps.
    }
    \label{fig:start_node_analysis}
\end{figure*}
\section{Interpretability Analysis of AM and POMO}\label{sec:pomo}

As noted in \Secref{sec:introduction}, both AM and POMO employ an asymmetrical architecture consisting of multiple encoder layers and a single decoder layer. Because the current decoding step is omitted from the encoder embeddings and processed exclusively by the lightweight decoder, we hypothesize that AM and POMO's construction strategy relies less on local geometric structures around the current node and more on a generalized global heuristic. To test this hypothesis, we examine the geometric properties of tours generated by these models and compare them against alternative solvers.

Following the methodology in \Secref{subsec:geometric-trajectory}, we track the angular displacement of the current node relative to the start node throughout tour construction. As shown in \Figref{fig:angle-progression}, a clear behavioral divergence emerges: while baseline methods dynamically adjust their angular direction without an explicit directional bias, \textbf{POMO exhibits a systematic tendency to construct tours in a clockwise manner while AM constructs in an anti-clockwise manner}.

Furthermore, \textbf{both of these models oscillate between different onion-layer depths with substantially higher frequency and intensity} than competing solvers (\Figref{fig:excursion-count-uniform}). This spatial disparity becomes increasingly pronounced on larger problem scales, helping explain why AM and POMO underperform relative to LEHD and L2C-Insert, both of which adapt more dynamically to local instance geometry (see \Secref{sec:lehd}).

This observation aligns with findings by \citet{huang2025rethinking}, who noted that lightweight decoder architectures yield less context-aware decoding decisions, ultimately limiting performance. To mitigate this constraint, light-decoder architectures frequently employ multi-start sampling at inference time, generating candidate tours from multiple initial nodes per instance and selecting the minimal-cost path. We argue that the efficacy of multi-start sampling stems directly from counteracting POMO's rigid geometric strategy. However, because test-time sampling provides only constrained flexibility compared to the innate adaptivity of greedy LEHD, the performance gap between the two models widens as instance sizes increase (\Appref{sec:behavioral-extended}).

This structural rigidity also accounts for AM and POMO's performance degradation on clustered distributions (\Figref{fig:tsp500_behavioral_comparison}). Because these models are trained exclusively on uniform point distributions, their learned heuristics function adequately on uniform data but fail to generalize to clustered instances, forcing unnecessary inter-cluster jumps that severely inflate total tour cost and hull violation. Interestingly, while POMO suffers a larger overall performance gap, its rigid clockwise progression better preserves the cyclic order of convex-hull nodes on near-uniform distributions, while degrading sharply on clustered and mixed distributions (\Figref{fig:tsp500_behavioral_comparison}). This also applies to AM.

To evaluate how training data distributions influence POMO's learned strategy, we retrained the model on clustered datasets from \citet{bi2023learning} and measured intra-cluster layer oscillations. As illustrated in \Figref{fig:excursion-count-cluster}, this retrained variant oscillates significantly more within individual clusters than baseline methods—and even exceeds the oscillation intensity of the original uniform-trained POMO. This shift demonstrates that the training distribution directly governs the learned heuristic: when trained on clustered instances, POMO restricts its layer oscillations within a cluster before transitioning to the next. This adaptation yields substantial performance gains across all evaluated instance distributions (\Figref{fig:tsp500_behavioral_comparison}), suggesting that diversifying training data distributions offers a promising avenue for enhancing NCO generalization more broadly. We trained AM and LEHD on clustered data as well, as depicted in \Appref{subsec:tsp-behavioral}.

Remarkably, the angular progressions of AM and POMO persist even in their cluster-trained variant, indicating that they apply their core strategy hierarchically within each cluster, effectively treating individual clusters as distinct subproblems. This reinforces \citet{bi2023learning}, demonstrating that light-decoder models operate similarly when solving full instances or decomposed subproblems.
\section{Interpretability Analysis of LEHD}\label{sec:lehd}
While POMO's routing strategy is relatively rigid, as discussed in \Secref{sec:pomo}, LEHD's strategy is considerably more adaptive to the underlying geometry of the nodes. Our analysis reveals that LEHD's navigation paradigm comprises two distinct components: a \textit{local navigation} mechanism that optimizes paths across immediate local neighborhoods, and a \textit{global navigation} mechanism that steers the local trajectory toward a broader, global target direction. Below, we dissect each component in detail.

\subsection{Local Navigation}\label{subsec:localnav}

How does LEHD select its immediate next node—and does it plan ahead when doing so? Because LEHD computes attention over all nodes during decoding, we first visualized its attention heatmaps. This revealed a striking pattern (see \Figref{fig:lehd-attention-heatmap}): at any given step, the current node attends heavily not only to the immediate next node but also to a sequence of nodes slated for addition in subsequent decoding steps. This suggests that LEHD selects its next node by explicitly accounting for its planned trajectory over a longer horizon, a strategy that shows some resemblance to Model Predictive Control (MPC) \citep{richalet1978model}. In MPC, a future path is optimized up to a fixed predictive horizon, but only the first control step is executed; looking ahead prevents short-sighted, purely greedy decisions.

To test whether future trajectory information is actively encoded in the decoder, we trained linear classifier probes on the latent representations from the final two layers. The full probing protocol is detailed in \Secref{subsec:probing}. We define the horizon as the number of decoding steps between the current node and the node being predicted.

Probing accuracies are shown in \Figref{fig:pomo_lehd_horizon_probe}, alongside AM and POMO as baselines. For POMO and AM, we probe candidate-wise products of decoder contexts and final encoder embeddings or logit keys, respectively. At horizon 1 (the immediate next node), all models achieve near-perfect accuracy, as expected: all three architectures use a linear projection over these candidate-specific representations to compute the next-node distribution. At longer horizons, however, a stark divergence emerges. LEHD predicts the node two steps ahead with approximately 80\% accuracy, indicating that substantial future-path information is explicitly retained in its latent space. AM and POMO, by contrast, degrade sharply, especially on larger instances, implying that such information is absent or heavily obscured in their representations. This gap persists up to horizons of 4–5 steps, beyond which LEHD's probe accuracy also decays, delineating the upper bound of its effective look-ahead horizon. To rule out the possibility of the probes predicting the nearest neighbor, in \Secref{subsec:lehd_non_nn_probe} we also evaluate them on instances where LEHD does not choose the nearest neighbor.

Probing establishes the \textit{presence} of future-path information, but not that the model uses it. To test causal relevance, we turn to the counterfactual steering described in \Secref{subsec:counterfactual-steering}. As shown in \Figref{fig:lehd-probe-steer}, slightly modifying the embeddings of nodes in the second most probable branch changes the model's preference toward choosing $a'_1$ as its immediate next action, without altering the embeddings of the immediate next nodes. Random-direction updates, or updates to nodes far along the horizon, leave LEHD's decision unchanged, consistent with the bounded effective horizon identified by the probes. We further assess potential off-manifold effects of these
interventions in \Appref{app:steering_off_manifold}. This provides causal evidence that LEHD relies on its encoded look-ahead trajectory to determine its local actions. A complementary behavioral causal experiment in \Appref{subsec:behavioral-causal} further examines POMO and LEHD and corroborates this conclusion.

\begin{table}[H]
\centering
\small
\begin{minipage}[t]{0.48\textwidth}
    \centering
    \small
    \setlength{\tabcolsep}{3pt}
    \renewcommand{\arraystretch}{1.08}
    \captionof{table}{
    TSP-100 rollout-order similarity comparison between LCS and Rev. LCS.
    }
    \label{tab:start-patching-lcs-n100}
    \begin{tabular}{lrr|rr}
    \toprule
    \multirow{2}{*}{Angle Range}
    & \multicolumn{2}{c|}{LEHD}
    & \multicolumn{2}{c}{POMO} \\
    \cmidrule(lr){2-3}\cmidrule(lr){4-5}
    & LCS
    & Rev.\ LCS
    & LCS
    & Rev.\ LCS \\
    \midrule
    $0^\circ$--$30^\circ$
    & \textbf{0.769} & 0.147
    & \textbf{0.963} & 0.030 \\

    $30^\circ$--$60^\circ$
    & \textbf{0.684} & 0.226
    & \textbf{0.963} & 0.033 \\

    $60^\circ$--$90^\circ$
    & \textbf{0.612} & 0.302
    & \textbf{0.965} & 0.036 \\

    $90^\circ$--$120^\circ$
    & \textbf{0.516} & 0.398
    & \textbf{0.966} & 0.039 \\

    $120^\circ$--$150^\circ$
    & 0.428 & \textbf{0.489}
    & \textbf{0.963} & 0.042 \\

    $150^\circ$--$180^\circ$
    & 0.391 & \textbf{0.544}
    & \textbf{0.963} & 0.052 \\
    \bottomrule
    \end{tabular}
\end{minipage}
\hfill
\begin{minipage}[t]{0.48\textwidth}
    \centering
    \small
    \setlength{\tabcolsep}{1pt}
    \renewcommand{\arraystretch}{1.08}
    \captionof{table}{Layer-wise cross-instance alignment of current, start, and random-node representations in LEHD on Uniform TSP-100. Enc. and Dec. denote encoder and decoder layers, respectively.}
    \label{tab:lehd-role-alignment}
    \begin{tabular}{lccc}
    \toprule
    \textbf{Layer} &
    \textbf{Current} &
    \textbf{Start} &
    \textbf{Random} \\
    \midrule
    Enc. 1 & $0.194 \pm 0.557$ & $0.188 \pm 0.559$ & $0.193 \pm 0.557$ \\
    Dec. 1 & $0.378 \pm 0.435$ & $\mathbf{0.457} \pm 0.305$ & $0.437 \pm 0.336$ \\
    Dec. 2 & $\mathbf{0.767} \pm 0.124$ & $0.595 \pm 0.256$ & $0.378 \pm 0.360$ \\
    Dec. 3 & $\mathbf{0.741} \pm 0.114$ & $0.630 \pm 0.222$ & $0.371 \pm 0.339$ \\
    Dec. 4 & $\mathbf{0.886} \pm 0.046$ & $0.715 \pm 0.175$ & $0.375 \pm 0.306$ \\
    Dec. 5 & $\mathbf{0.732} \pm 0.109$ & $0.691 \pm 0.167$ & $0.373 \pm 0.269$ \\
    \bottomrule
    \end{tabular}
\end{minipage}
\end{table}

\subsection{Global Navigation}
\label{subsec:globalnav}
While local navigation governs immediate look-ahead dependencies, LEHD also relies on a global navigation component that steers the route toward a broader target direction. To isolate the role of the start node in this mechanism, we first compare it with the current node using the mean-ablation intervention described in \Secref{subsubsec:mean-ablation}. As shown in \Figref{fig:mean_ablation}, ablating the current node produces a much larger drop in the clean next-node probability than ablating the start node. This indicates that immediate next-node selection is governed primarily by the current local state, whereas the start node has a weaker direct effect on local transitions and may instead support higher-level guidance.

We examine this role more directly using the activation-transfer intervention introduced in \Secref{subsubsec:activation-transfer}, where the original start-node representation \(S\) is replaced by that of a donor node \(S'\). As shown in \Figref{fig:start_patching_plot}, \textit{maximum angle without distance constraints} produces the largest drop in the clean next-node probability, followed by \textit{maximum angle under similar distance}, while \textit{similar angle with different distance} produces the smallest effect. The first two conditions preserve a large angular displacement under different distance constraints, whereas the third substantially changes distance while approximately preserving direction. This ordering suggests that the intervention effect is more strongly associated with the angular displacement \(\angle(S,C,S')\), defined by the original start node, the current node, and the donor node, than with the distance between \(S\) and \(S'\). In contrast, the corresponding POMO curves remain comparatively close to zero and exhibit little separation across the three donor conditions. Thus, POMO's immediate decisions appear largely insensitive to the directional information introduced through the patched start representation, whereas LEHD responds strongly to changes in this direction. The corresponding results for AM, which selects its starting node
through its learned policy, are discussed in
\Secref{subsec:am-start-navigation}.

The qualitative example in \Figref{fig:trajectory_instance} illustrates the resulting reorientation. In this instance, the patched start node lies at an angle of nearly \(180^\circ\) relative to the original start node, and the generated route changes its overall direction so that the remaining trajectory approaches the patched anchor through a shorter completion path. To quantify this effect, \Tabref{tab:start-patching-lcs-n100} compares the ordering of the unvisited nodes in the clean and patched rollouts. LCS is the normalized longest common subsequence between the clean rollout order and the patched rollout order, and therefore measures how much of the original forward ordering is preserved. Rev.\ LCS instead compares the patched rollout order with the reversed clean rollout, and measures whether the intervention causes the remaining nodes to be visited in the opposite direction.

For LEHD, increasing \(\angle(S,C,S')\) consistently decreases LCS and increases Rev.\ LCS. At the largest angular displacements, reversed agreement exceeds forward agreement, indicating a substantial reordering of the future selection sequence toward the reverse direction. In contrast, POMO maintains relatively high LCS and low Rev.\ LCS across the full range of angular interventions, suggesting that its rollout order remains comparatively stable even under large directional perturbations. Together, these results suggest that the start node provides global directional guidance in LEHD, but has much less influence in POMO.

\subsection{Latent Space}
\label{subsec:lehd-latent}

To examine whether LEHD organizes node representations according to their functional roles, we apply the cross-instance alignment analysis described in \Secref{subsec:node-role-alignment}. As shown in Table~\ref{tab:lehd-role-alignment}, the current-node representation exhibits high cosine similarity across instances from decoder layer 2 onward. Although the current node changes at every decoding step and occupies different geometric locations across instances, its representation remains strongly aligned at the same step. This suggests that LEHD maps functionally equivalent current nodes into a shared latent region, potentially providing a consistent reference for subsequent decisions. The start node, by contrast, remains fixed throughout the rollout, yet its alignment increases more gradually and peaks in deeper decoder layers, consistent with a broader navigational role becoming more strongly expressed later in the computation. Randomly selected unvisited nodes show larger variability and less consistent alignment. Overall, these results suggest that LEHD progressively organizes its latent space around distinct task-specific node roles. Additional visualizations are provided in \Appref{sec:visualization}.
\section{Conclusion and Future Work}\label{sec:conclusion}
In this work, we conducted an extensive and diverse set of experiments to provide a comprehensive understanding of the strategies underlying three of the most well-known neural NCO solvers for routing problems. Our results reveal that these models employ distinct strategies to solve TSP and CVRP instances. These insights, and the methods used to derive them, can inform the design of better solvers and guide future work aimed at deepening our understanding of these models.

Several promising future directions emerge from this study. First, training on more diverse data distributions could improve robustness and generalization. Second, LEHD could be trained to predict multiple nodes along the path at once instead of only the next node, enabling it to plan further ahead and potentially strengthening its routing capability. Third, increasing the number of decoder layers relative to encoder layers appears to increase model flexibility and improve generalization. While this work focuses on three representative NCO solvers, extending the analysis to other solvers is another interesting direction.
\section*{AI Use Statement}\label{sec:ai_statement}

In this work, we used generative AI tools for implementing methods (specifically, assisting in the implementation of some experiments). We have not used generative AI tools for generating synthetic data sets, developing theoretical models or conceptual frameworks, formulating or proving mathematical claims, proposing or refining hypotheses, designing research methodology or experiments, supporting qualitative or thematic data analysis, interpreting results, or assisting with translation or dataset cleaning and reformatting. Additionally, we used generative AI tools for creating or editing software code (primarily for initial code scaffolding, such as generating plots or loading raw data) and for editing the paper to improve readability (proofreading and polishing the text of this manuscript). We have reviewed all AI-assisted work. Specifically, all code generated with LLM assistance was rigorously reviewed, tested, and validated by multiple co-authors to ensure its correctness and integration into our codebase, and all AI-assisted text edits were reviewed by the authors for accuracy and intent. We take responsibility for the final content of this work, including text, claims or artifacts produced with the aid of generative AI.

\bibliography{main}
\bibliographystyle{main}

\clearpage

\appendix
\section{Preliminaries}\label{sec:preliminaries}

\subsection{Traveling Salesman Problem}\label{subsec:tsp}
Given a complete graph $G=(V,E)$ with $n$ nodes and edge costs $c_{ij}$, the TSP seeks a minimum-cost Hamiltonian cycle visiting each node exactly once. Formally, we seek a permutation $\pi=[\pi_0,\pi_1,\ldots,\pi_{n-1}]$ that minimizes:
\begin{equation}
\min_{\pi} \left( \sum_{i=0}^{n-2} c_{\pi_i,\pi_{i+1}} + c_{\pi_{n-1},\pi_0} \right).
\end{equation}

\subsection{Capacitated Vehicle Routing Problem}
\label{subsec:cvrp}
In the CVRP, a vehicle with capacity $Q$ must service customer demands $d_i>0$ from a central depot $v_0$. The goal is to find a set of routes $\mathcal{R}$ minimizing total travel cost while respecting capacity constraints:
\begin{equation}
\begin{aligned}
\min_{\mathcal{R}} \quad &
\sum_{\mathcal{R}_k \in \mathcal{R}}
\sum_{t=0}^{|\mathcal{R}_k|-2}
c_{r_t^k,r_{t+1}^k},\\
\text{s.t.} \quad &
\sum_{t=1}^{|\mathcal{R}_k|-2}
d_{r_t^k} \le Q,
\quad \forall \mathcal{R}_k \in \mathcal{R}.
\end{aligned}
\end{equation}
Each route $\mathcal{R}_k=[r_0^k,r_1^k,\ldots,r_{|\mathcal{R}_k|-1}^k]$ starts and ends at the depot ($r_0^k = r_{|\mathcal{R}_k|-1}^k = v_0$), and every customer is visited exactly once.

\subsection{Onion Decomposition}\label{subsec:onion-decomposition}
Onion decomposition recursively removes the convex hull of the remaining points, partitioning them into nested layers. A node has onion depth \(d(v)=k\) if it is removed at peeling iteration \(k\); outer-hull nodes have depth zero, while larger values indicate deeper layers.

Let \(\pi_t\) denote the node selected at decoding step \(t\). We normalize its depth as

\begin{equation}
\rho_t
=
\frac{
d\left(\pi_t\right)
}{
\max\left(
1,
\max_{v\in\mathcal{V}} d(v)
\right)
}.
\label{eq:normalized-onion-depth}
\end{equation}
\clearpage
\section{Related Work}\label{sec:relatedwork}

\subsection{NCO Solvers}\label{subsec:nco_solvers}

\subsubsection{Pointer Networks} The paradigm of using deep neural networks to learn heuristics for solving NP-hard routing problems was pioneered by Pointer Networks \citep{vinyals2017pointer}. Unlike traditional sequence-to-sequence models that are constrained by a fixed output vocabulary, Pointer Networks employ a modified attention mechanism as a dynamic pointing layer, enabling the model to select outputs directly from variable-length input sequences.

\subsubsection{AM} While Pointer Networks were originally implemented using recurrent architectures such as LSTMs \citep{Hochreiter1997long}, the Attention Model (AM) \citep{kool2019attention} adopted a Transformer-based architecture. AM achieves permutation invariance, allowing the model to treat the input graph as an unordered spatial set. To eliminate the need for ground-truth optimal solutions during training, AM introduced the use of reinforcement learning through the REINFORCE algorithm \citep{williams1992reinforce}, together with a rollout baseline to reduce gradient variance.

\subsubsection{POMO} Although AM demonstrated the effectiveness of Transformer-based routing solvers, its training process still suffered from high variance. Policy Optimization with Multiple Optima (POMO) \citep{kwon2021pomo} addressed this limitation by exploiting the rotational and structural symmetries inherent in cyclic routing problems such as the TSP. Instead of constructing a single trajectory, POMO generates $N$ trajectories for a given graph instance, each initialized from a different starting node.

\subsubsection{LEHD} A persistent limitation of both AM and POMO is their difficulty in generalizing to graph sizes substantially larger than those encountered during training. Unlike POMO, LEHD \citep{luo2024neural} adopts a supervised learning paradigm. It employs a Transformer-based architecture consisting of a one-layer encoder and a six-layer decoder. At each decoding step, the model receives the starting node, the current node, and the set of unvisited nodes, and predicts the next node to append to the partial route.

\subsubsection{L2C-Insert} Both POMO and LEHD are construction heuristics that generate solutions through successive node appending. Aiming to overcome the limitations of appending-based strategies, \citet{luo2025learning} proposed L2C-Insert, an insertion-based solver. The method consists of a node-selection phase, in which a candidate node is chosen based on its proximity to the previously selected node, followed by an insertion phase. During the insertion phase, a Transformer encoder-decoder architecture determines the optimal position of the selected node within the current partial tour, thereby incrementally expanding the solution.

\subsection{Mechanistic Interpretability}\label{subsec:mechanistic_interpretability}

Mechanistic Interpretability (MI) seeks to uncover the internal computations performed by neural networks by identifying the information encoded in their representations and establishing causal relationships between model components and observed behavior.

\subsubsection{Probing} Probing is one of the most widely used approaches for analyzing neural representations. A probe is typically a simple linear classifier or regressor trained to predict a target property from a model's hidden activations. Strong probe performance suggests that information relevant to the target property is encoded in the examined representation \citep{alain2018understanding}. Consequently, probing has become a standard tool for investigating the information accessible at different layers of a neural network \citep{belinkov2021probing}.

\subsubsection{Activation Patching} While probing can reveal whether information is present in a representation, it cannot determine whether that information is causally used by the model. Activation patching addresses this limitation by replacing activations from one input with the corresponding activations from another and measuring the resulting change in model behavior. By analyzing the effects of such interventions, activation patching can identify components that are causally responsible for specific computations \citep{meng2022locating,wang2022interpretability,zhang2023towards,geiger2025causal}. The technique has been successfully applied to uncover computational circuits and information flow in Transformer-based models \citep{heimersheim2024use}.

\subsubsection{Look-Ahead and Planning in Sequential Decision-Making Models}
Outside of NLP, a closely related line of MI work asks whether networks trained end-to-end on sequential decision-making tasks represent their own future actions, and whether these representations are causally used rather than merely correlated with behavior. \citet{jenner2024evidence} combine linear probing with activation patching on Leela Chess Zero, the strongest open-source chess engine, and find that its policy network linearly encodes the optimal move several turns ahead of the current position, and that these representations are causally necessary for its output in certain board states. In a parallel line of work on a different planning domain, \citet{taufeeque2025planning} train linear probes that decode a recurrent network's future actions roughly 50 steps in advance while it plays Sokoban, and show via intervention on the hidden state that these probed representations causally steer the agent's subsequent behavior; \citet{taufeeque2026path} extend this analysis into a full circuit-level account, localizing directional ``path channels'' that implement a bidirectional, plan-extending search. Both lines of work establish the same two-step recipe that we adopt: probing to establish that a representation of a future decision exists, followed by causal intervention to establish that the model actually relies on it. Our contribution is to bring this recipe to bear on NCO routing solvers, where a variable-size, permutation-sensitive candidate set and the absence of a fixed board or grid structure require a different probe formulation --- candidate-ranking over the feasible action set rather than a fixed square- or move-indexed classification target --- than either the chess or Sokoban settings.

\subsection{Interpretability of NCO Solvers}\label{subsec:interpretability_nco}

Despite the rapid progress of neural combinatorial optimization, the internal mechanisms underlying these solvers remain largely unexplored. One of the first studies to investigate this question is the work of \citet{zhang2025probing}. Using probing techniques, they showed that Euclidean distance information is encoded in the internal representations of AM, POMO, and LEHD. Furthermore, their analysis suggested that these models do not behave purely myopically when constructing solutions: their myopia-avoidance probe is framed as a single-step, binary classification between the globally optimal edge and the locally greedy (nearest-neighbor) edge at the current decision, evaluated on the same two backbones we study. This probe does not test whether the model encodes a specific candidate action at longer horizons, nor whether any such representation is causally used, both of which are the focus of our Future-Action Planning Probes and the causal interventions. Also by examining the weights of the probe they found that there exists two dimensions in the latent space of LEHD which have high importance in its node selection.

More recently, \citet{narad2025mechanistic} applied sparse autoencoders (SAEs) to a pointer network-style model \citep{vinyals2017pointer} and found evidence that the learned representations capture interpretable geometric concepts, including boundary detection and spatial clustering; the authors explicitly identify causal circuit analysis via activation patching as future work rather than something their study performs.

Complementary to these mechanistic approaches, \citet{Kikuta_2024} proposed RouteExplainer, a post-hoc explanation framework for VRP solutions. RouteExplainer quantifies the influence of individual edges on the generated route, introduces a pipeline for generating counterfactual explanations, and leverages large language models (LLMs) to enhance the interpretability of the resulting explanations.

Taken together, existing interpretability studies of NCO solvers are exclusively correlational --- probing \citep{zhang2025probing}, sparse autoencoders \citep{narad2025mechanistic}, post-hoc explanation \citep{Kikuta_2024}, --- and none establishes a causal link between a specific internal representation and a specific routing decision. To the best of our knowledge, the activation-patching experiments in Section~4.4 are the first causal account of decision-making mechanisms in NCO routing solvers specifically, extending the look-ahead-probing-and-patching paradigm established outside NLP (above) to this domain for the first time.
\clearpage
\section{Extended Behavioral Results}\label{sec:behavioral-extended}
This section extends the behavioral evaluation presented in the main paper.
We report detailed results for TSP and CVRP across multiple problem sizes
and seven distribution shifts, and subsequently examine the rotational
sensitivity of the models. These experiments provide a broader comparison
of solution quality, geometric consistency, scale generalization, and
robustness under changes in instance geometry.

\subsection{TSP Behavioral Results}\label{subsec:tsp-behavioral}

We compare L2C-Insert, LEHD, POMO, AM, and cluster-trained variants of
POMO and LEHD across TSP sizes from 20 to 1000 and seven distribution
shifts. We report the optimality gap relative to Concorde, the average
number of edge crossings, and the convex-hull order violation rate. An
optimal Euclidean TSP tour contains no edge crossings and visits convex-hull
vertices in their cyclic order; therefore, these two metrics capture
complementary forms of geometric inconsistency. Lower values are better,
and the best result for each metric, including ties, is shown in bold.

Tables~\ref{tab:tsp-20-four-model-comparison}--%
\ref{tab:tsp-1000-four-model-comparison} reveal a clear separation between
tour quality and geometric regularity. At smaller problem sizes,
L2C-Insert and standard LEHD are generally competitive, whereas POMO and
AM tend to exhibit larger optimality gaps. As the problem size increases,
however, LEHD-Cluster becomes increasingly strong and consistently
outperforms standard LEHD. This effect is especially pronounced on
TSP-500 and TSP-1000, where LEHD-Cluster achieves the best or near-best
optimality gaps across the tested distributions.

Interestingly, this improvement appears to stem from cluster-based
training rather than from the architecture alone. LEHD-Cluster generalizes
substantially better than standard LEHD as the problem size grows,
including on unseen distributions. One plausible explanation is that
cluster training encourages a more decompositional strategy, where the
model solves spatially coherent regions as smaller subproblems before
connecting them. We view this as a behavioral hypothesis rather than
direct mechanistic evidence.

Cluster training also substantially improves the large-scale generalization
of POMO, reducing its optimality gap across all seven distributions on
TSP-500 and TSP-1000. Together with the LEHD-Cluster results, this
suggests that clustered training can benefit both architectures by
improving how they exploit spatial structure.

The geometric metrics provide a complementary view. POMO and
POMO-Cluster can exhibit relatively low convex-hull violation rates despite
substantially larger optimality gaps, indicating that preserving coarse
geometric structure is not sufficient for producing a low-cost tour. AM
shows a similar separation in several settings, where geometrically regular
behavior does not necessarily translate into strong tour quality. Overall,
the results suggest that large-scale generalization depends not only on
preserving geometric constraints, but also on learning a solution strategy
that can organize and coordinate decisions across increasingly large
instances.


\begin{table*}[ht!]
\centering
\scriptsize
\setlength{\tabcolsep}{2.0pt}
\renewcommand{\arraystretch}{1.08}
\caption{Solution quality and geometric consistency under greedy decoding
on TSP-20 across seven test distributions.}
\label{tab:tsp-20-four-model-comparison}
\resizebox{\textwidth}{!}{%
\begin{tabular}{@{}l*{18}{c}@{}}
\toprule
\multirow{2}{*}{Distribution}
& \multicolumn{3}{c}{L2C-Insert}
& \multicolumn{3}{c}{LEHD}
& \multicolumn{3}{c}{POMO}
& \multicolumn{3}{c}{AM}
& \multicolumn{3}{c}{POMO (Cluster-Trained)}
& \multicolumn{3}{c}{LEHD (Cluster-Trained)} \\
\cmidrule(lr){2-4}
\cmidrule(lr){5-7}
\cmidrule(lr){8-10}
\cmidrule(lr){11-13}
\cmidrule(lr){14-16}
\cmidrule(lr){17-19}
& Gap (\%) & Cross. & Hull Viol. (\%)
& Gap (\%) & Cross. & Hull Viol. (\%)
& Gap (\%) & Cross. & Hull Viol. (\%)
& Gap (\%) & Cross. & Hull Viol. (\%)
& Gap (\%) & Cross. & Hull Viol. (\%)
& Gap (\%) & Cross. & Hull Viol. (\%) \\
\midrule

Clustered
& 0.55 & 0.06 & 3.83
& 0.49 & 0.07 & 4.86
& 9.41 & 0.49 & 8.22
& 26.68 & 2.99 & 51.24
& 3.37 & 0.18 & 3.00
& \textbf{0.19} & \textbf{0.01} & \textbf{0.95} \\

Expansion
& 3.55 & 0.39 & 20.96
& 3.04 & 0.30 & 13.78
& 6.51 & 0.50 & 8.36
& 17.72 & 2.04 & 26.15
& 4.48 & 0.23 & \textbf{2.00}
& \textbf{0.93} & \textbf{0.07} & 2.97 \\

Explosion
& 2.67 & 0.26 & 9.99
& 1.13 & 0.08 & 3.66
& 5.62 & 0.34 & 6.11
& 9.70 & 0.72 & 13.48
& 3.83 & 0.37 & 5.00
& \textbf{0.59} & \textbf{0.03} & \textbf{0.90} \\

Grid
& 2.22 & 0.21 & 8.90
& 0.96 & 0.07 & 3.16
& 5.33 & 0.31 & 5.75
& 8.12 & 0.52 & 10.34
& 3.63 & 0.16 & 2.00
& \textbf{0.61} & \textbf{0.02} & \textbf{0.77} \\

Implosion
& 2.04 & 0.20 & 8.48
& 0.95 & 0.07 & 3.24
& 5.27 & 0.30 & 5.39
& 8.36 & 0.54 & 10.15
& 3.68 & 0.24 & 4.00
& \textbf{0.62} & \textbf{0.02} & \textbf{0.77} \\

Mixed
& 2.42 & 0.25 & 11.08
& 1.44 & 0.13 & 5.44
& 8.34 & 0.66 & 9.39
& 12.91 & 1.22 & 18.56
& 6.25 & 0.43 & 4.00
& \textbf{0.98} & \textbf{0.07} & \textbf{2.29} \\

Uniform
& 2.18 & 0.21 & 8.89
& 0.94 & 0.07 & 3.08
& 5.27 & 0.29 & 5.13
& 8.20 & 0.53 & 10.28
& 3.60 & 0.18 & 2.00
& \textbf{0.61} & \textbf{0.03} & \textbf{1.01} \\

\bottomrule
\end{tabular}%
}
\end{table*}


\begin{table*}[ht!]
\centering
\scriptsize
\setlength{\tabcolsep}{2.0pt}
\renewcommand{\arraystretch}{1.08}
\caption{Solution quality and geometric consistency under greedy decoding
on TSP-50 across seven test distributions.}
\label{tab:tsp-50-four-model-comparison}
\resizebox{\textwidth}{!}{%
\begin{tabular}{@{}l*{18}{c}@{}}
\toprule
\multirow{2}{*}{Distribution}
& \multicolumn{3}{c}{L2C-Insert}
& \multicolumn{3}{c}{LEHD}
& \multicolumn{3}{c}{POMO}
& \multicolumn{3}{c}{AM}
& \multicolumn{3}{c}{POMO (Cluster-Trained)}
& \multicolumn{3}{c}{LEHD (Cluster-Trained)} \\
\cmidrule(lr){2-4}
\cmidrule(lr){5-7}
\cmidrule(lr){8-10}
\cmidrule(lr){11-13}
\cmidrule(lr){14-16}
\cmidrule(lr){17-19}
& Gap (\%) & Cross. & Hull Viol. (\%)
& Gap (\%) & Cross. & Hull Viol. (\%)
& Gap (\%) & Cross. & Hull Viol. (\%)
& Gap (\%) & Cross. & Hull Viol. (\%)
& Gap (\%) & Cross. & Hull Viol. (\%)
& Gap (\%) & Cross. & Hull Viol. (\%) \\
\midrule

Clustered
& 0.65 & 0.05 & 2.24
& 0.91 & 0.14 & 7.10
& 5.26 & 0.27 & 2.31
& 33.42 & 6.57 & 41.24
& 1.72 & 0.04 & \textbf{0.00}
& \textbf{0.26} & \textbf{0.02} & 0.42 \\

Expansion
& 2.61 & 0.43 & 17.21
& 4.21 & 0.67 & 28.55
& 2.69 & 0.31 & 3.26
& 13.38 & 2.76 & 11.25
& 3.12 & 0.21 & \textbf{2.00}
& \textbf{0.94} & \textbf{0.10} & 4.00 \\

Explosion
& \textbf{0.36} & \textbf{0.02} & 0.75
& 0.55 & 0.05 & 1.41
& 1.37 & 0.09 & \textbf{0.74}
& 5.71 & 0.24 & 1.93
& 2.19 & 0.11 & 1.00
& 0.66 & 0.05 & 0.95 \\

Grid
& \textbf{0.39} & \textbf{0.02} & 0.51
& 0.50 & 0.03 & 0.74
& 0.85 & \textbf{0.02} & 0.07
& 4.36 & 0.12 & 0.57
& 2.34 & 0.05 & \textbf{0.00}
& 0.73 & 0.04 & 0.97 \\

Implosion
& \textbf{0.38} & \textbf{0.02} & 0.50
& 0.50 & 0.03 & 0.84
& 0.89 & 0.03 & 0.09
& 4.54 & 0.13 & 0.73
& 2.28 & 0.04 & \textbf{0.00}
& 0.69 & 0.04 & 0.92 \\

Mixed
& \textbf{0.74} & \textbf{0.05} & 1.34
& 0.89 & 0.13 & 2.66
& 3.31 & 0.41 & 2.85
& 6.81 & 0.46 & 3.13
& 4.25 & 0.19 & \textbf{0.00}
& 1.15 & 0.13 & 2.31 \\

Uniform
& \textbf{0.37} & \textbf{0.01} & 0.46
& 0.51 & 0.03 & 0.76
& 0.87 & 0.03 & \textbf{0.07}
& 4.34 & 0.11 & 0.56
& 2.34 & 0.10 & 1.00
& 0.72 & 0.04 & 1.06 \\

\bottomrule
\end{tabular}%
}
\end{table*}


\begin{table*}[ht!]
\centering
\scriptsize
\setlength{\tabcolsep}{2.0pt}
\renewcommand{\arraystretch}{1.08}
\caption{Solution quality and geometric consistency under greedy decoding
on TSP-100 across seven test distributions.}
\label{tab:tsp-100-four-model-comparison}
\resizebox{\textwidth}{!}{%
\begin{tabular}{@{}l*{18}{c}@{}}
\toprule
\multirow{2}{*}{Distribution}
& \multicolumn{3}{c}{L2C-Insert}
& \multicolumn{3}{c}{LEHD}
& \multicolumn{3}{c}{POMO}
& \multicolumn{3}{c}{AM}
& \multicolumn{3}{c}{POMO (Cluster-Trained)}
& \multicolumn{3}{c}{LEHD (Cluster-Trained)} \\
\cmidrule(lr){2-4}
\cmidrule(lr){5-7}
\cmidrule(lr){8-10}
\cmidrule(lr){11-13}
\cmidrule(lr){14-16}
\cmidrule(lr){17-19}
& Gap (\%) & Cross. & Hull Viol. (\%)
& Gap (\%) & Cross. & Hull Viol. (\%)
& Gap (\%) & Cross. & Hull Viol. (\%)
& Gap (\%) & Cross. & Hull Viol. (\%)
& Gap (\%) & Cross. & Hull Viol. (\%)
& Gap (\%) & Cross. & Hull Viol. (\%) \\
\midrule

Clustered
& 1.06 & 0.13 & 3.17
& 1.49 & 0.36 & 12.71
& 7.00 & 1.19 & 8.26
& 41.40 & 13.61 & 39.81
& 2.61 & 0.23 & \textbf{0.00}
& \textbf{0.45} & \textbf{0.04} & 0.70 \\

Expansion
& 2.55 & 0.62 & 20.14
& 5.16 & 1.51 & 49.94
& 4.59 & 1.29 & 9.99
& 12.63 & 3.29 & 7.38
& 3.86 & 0.24 & \textbf{1.00}
& \textbf{1.09} & \textbf{0.16} & 5.07 \\

Explosion
& \textbf{0.45} & \textbf{0.02} & \textbf{0.37}
& 0.63 & 0.08 & 1.37
& 2.12 & 0.46 & 3.49
& 6.55 & 0.45 & 0.47
& 2.75 & 0.19 & 1.00
& 0.75 & 0.09 & 1.28 \\

Grid
& \textbf{0.47} & \textbf{0.01} & 0.19
& 0.56 & 0.03 & 0.42
& 0.90 & 0.07 & 0.07
& 4.32 & 0.19 & 0.03
& 3.20 & 0.24 & \textbf{0.00}
& 0.85 & 0.06 & 1.19 \\

Implosion
& \textbf{0.46} & \textbf{0.02} & 0.13
& 0.59 & 0.04 & 0.70
& 0.87 & 0.07 & 0.04
& 4.50 & 0.22 & 0.07
& 3.06 & 0.18 & \textbf{0.00}
& 0.83 & 0.06 & 0.83 \\

Mixed
& \textbf{0.86} & \textbf{0.09} & 0.95
& 1.03 & 0.24 & 3.96
& 4.77 & 1.66 & 7.96
& 7.24 & 0.59 & \textbf{0.65}
& 4.66 & 0.82 & 1.00
& 1.41 & 0.26 & 3.87 \\

Uniform
& \textbf{0.46} & \textbf{0.02} & 0.30
& 0.57 & 0.03 & 0.63
& 0.85 & 0.06 & 0.08
& 4.35 & 0.20 & \textbf{0.03}
& 3.18 & 0.16 & 1.00
& 0.85 & 0.07 & 1.34 \\

\bottomrule
\end{tabular}%
}
\end{table*}


\begin{table*}[ht!]
\centering
\scriptsize
\setlength{\tabcolsep}{2.0pt}
\renewcommand{\arraystretch}{1.08}
\caption{Solution quality and geometric consistency under greedy decoding
on TSP-200 across seven test distributions.}
\label{tab:tsp-200-four-model-comparison}
\resizebox{\textwidth}{!}{%
\begin{tabular}{@{}l*{18}{c}@{}}
\toprule
\multirow{2}{*}{Distribution}
& \multicolumn{3}{c}{L2C-Insert}
& \multicolumn{3}{c}{LEHD}
& \multicolumn{3}{c}{POMO}
& \multicolumn{3}{c}{AM}
& \multicolumn{3}{c}{POMO (Cluster-Trained)}
& \multicolumn{3}{c}{LEHD (Cluster-Trained)} \\
\cmidrule(lr){2-4}
\cmidrule(lr){5-7}
\cmidrule(lr){8-10}
\cmidrule(lr){11-13}
\cmidrule(lr){14-16}
\cmidrule(lr){17-19}
& Gap (\%) & Cross. & Hull Viol. (\%)
& Gap (\%) & Cross. & Hull Viol. (\%)
& Gap (\%) & Cross. & Hull Viol. (\%)
& Gap (\%) & Cross. & Hull Viol. (\%)
& Gap (\%) & Cross. & Hull Viol. (\%)
& Gap (\%) & Cross. & Hull Viol. (\%) \\
\midrule

Clustered
& 2.39 & 0.48 & 6.74
& 2.61 & 0.71 & 18.70
& 14.59 & 7.83 & 22.10
& 33.81 & 15.57 & 21.70
& 7.72 & 1.99 & \textbf{1.00}
& \textbf{1.36} & \textbf{0.10} & 2.06 \\

Expansion
& \textbf{0.86} & \textbf{0.10} & 0.30
& 1.52 & 0.33 & 0.92
& 3.61 & 0.71 & 0.02
& 7.07 & 0.46 & \textbf{0.00}
& 7.42 & 1.01 & \textbf{0.00}
& 1.29 & 0.16 & 0.68 \\

Explosion
& \textbf{0.92} & \textbf{0.08} & 0.62
& 1.08 & 0.09 & 0.92
& 4.03 & 0.67 & 0.14
& 7.16 & 0.43 & \textbf{0.08}
& 7.97 & 0.90 & 3.00
& 1.41 & 0.14 & 2.02 \\

Grid
& 1.01 & 0.09 & 2.32
& \textbf{0.90} & \textbf{0.08} & 1.86
& 5.23 & 1.39 & 0.52
& 8.64 & 0.53 & \textbf{0.04}
& 8.92 & 1.64 & 1.00
& 1.10 & 0.16 & 2.26 \\

Implosion
& 1.41 & 0.20 & 3.22
& 1.53 & 0.34 & 9.60
& 4.71 & 1.57 & \textbf{0.74}
& 17.76 & 2.14 & 0.94
& 7.73 & 1.66 & 2.00
& \textbf{0.89} & \textbf{0.10} & 1.98 \\

Mixed
& 1.24 & \textbf{0.21} & 2.96
& \textbf{1.16} & 0.27 & 5.16
& 7.70 & 4.94 & 7.38
& 9.46 & 1.23 & 0.26
& 8.39 & 2.54 & \textbf{0.00}
& 1.29 & 0.30 & 4.34 \\

Uniform
& 0.94 & \textbf{0.09} & 1.28
& \textbf{0.88} & \textbf{0.09} & 2.04
& 4.22 & 1.38 & 0.52
& 8.18 & 0.88 & \textbf{0.04}
& 7.83 & 1.62 & 1.00
& 1.02 & 0.14 & 2.54 \\

\bottomrule
\end{tabular}%
}
\end{table*}


\begin{table*}[ht!]
\centering
\scriptsize
\setlength{\tabcolsep}{2.0pt}
\renewcommand{\arraystretch}{1.08}
\caption{Solution quality and geometric consistency under greedy decoding
on TSP-500 across seven test distributions.}
\label{tab:tsp-500-four-model-comparison}
\resizebox{\textwidth}{!}{%
\begin{tabular}{@{}l*{18}{c}@{}}
\toprule
\multirow{2}{*}{Distribution}
& \multicolumn{3}{c}{L2C-Insert}
& \multicolumn{3}{c}{LEHD}
& \multicolumn{3}{c}{POMO}
& \multicolumn{3}{c}{AM}
& \multicolumn{3}{c}{POMO (Cluster-Trained)}
& \multicolumn{3}{c}{LEHD (Cluster-Trained)} \\
\cmidrule(lr){2-4}
\cmidrule(lr){5-7}
\cmidrule(lr){8-10}
\cmidrule(lr){11-13}
\cmidrule(lr){14-16}
\cmidrule(lr){17-19}
& Gap (\%) & Cross. & Hull Viol. (\%)
& Gap (\%) & Cross. & Hull Viol. (\%)
& Gap (\%) & Cross. & Hull Viol. (\%)
& Gap (\%) & Cross. & Hull Viol. (\%)
& Gap (\%) & Cross. & Hull Viol. (\%)
& Gap (\%) & Cross. & Hull Viol. (\%) \\
\midrule

Clustered
& 5.58 & 6.45 & 41.06
& 5.41 & 3.86 & 47.98
& 43.54 & 70.07 & 59.50
& 51.78 & 50.90 & 25.04
& 24.71 & 15.83 & \textbf{3.00}
& \textbf{1.91} & \textbf{0.69} & 14.36 \\

Expansion
& 2.29 & 1.00 & 3.46
& 2.72 & 1.52 & 3.52
& 34.72 & 27.80 & 1.72
& 18.27 & 2.98 & \textbf{0.00}
& 24.14 & 10.49 & \textbf{0.00}
& \textbf{1.75} & \textbf{0.61} & 2.24 \\

Explosion
& 2.16 & 0.77 & 7.26
& 2.05 & 0.76 & 5.46
& 33.14 & 22.83 & 3.16
& 18.78 & 3.12 & \textbf{0.52}
& 24.92 & 9.31 & 2.00
& \textbf{1.90} & \textbf{0.60} & 5.60 \\

Grid
& 2.29 & 1.35 & 17.86
& 1.77 & 0.87 & 19.22
& 26.94 & 30.47 & 8.88
& 21.14 & 2.79 & \textbf{0.10}
& 22.19 & 9.16 & 3.00
& \textbf{1.34} & \textbf{0.61} & 12.78 \\

Implosion
& 4.30 & 3.69 & 34.22
& 3.80 & 2.19 & 35.60
& 32.98 & 49.08 & 16.46
& 34.14 & 6.57 & \textbf{1.36}
& 24.21 & 14.69 & 5.00
& \textbf{1.27} & \textbf{0.51} & 13.12 \\

Mixed
& 2.89 & 2.61 & 19.62
& 2.30 & 1.61 & 19.46
& 35.67 & 62.64 & 34.04
& 21.33 & 4.72 & \textbf{1.04}
& 25.15 & 16.29 & 8.00
& \textbf{1.73} & \textbf{1.27} & 17.30 \\

Uniform
& 2.19 & 1.18 & 15.50
& 1.62 & 0.75 & 15.26
& 34.01 & 50.44 & 16.34
& 20.95 & 4.75 & \textbf{0.96}
& 24.56 & 15.81 & 3.00
& \textbf{1.36} & \textbf{0.60} & 11.82 \\

\bottomrule
\end{tabular}%
}
\end{table*}


\begin{table*}[ht!]
\centering
\scriptsize
\setlength{\tabcolsep}{2.0pt}
\renewcommand{\arraystretch}{1.08}
\caption{Solution quality and geometric consistency under greedy decoding
on TSP-1000 across seven test distributions.}
\label{tab:tsp-1000-four-model-comparison}
\resizebox{\textwidth}{!}{%
\begin{tabular}{@{}l*{18}{c}@{}}
\toprule
\multirow{2}{*}{Distribution}
& \multicolumn{3}{c}{L2C-Insert}
& \multicolumn{3}{c}{LEHD}
& \multicolumn{3}{c}{POMO}
& \multicolumn{3}{c}{AM}
& \multicolumn{3}{c}{POMO (Cluster-Trained)}
& \multicolumn{3}{c}{LEHD (Cluster-Trained)} \\
\cmidrule(lr){2-4}
\cmidrule(lr){5-7}
\cmidrule(lr){8-10}
\cmidrule(lr){11-13}
\cmidrule(lr){14-16}
\cmidrule(lr){17-19}
& Gap (\%) & Cross. & Hull Viol. (\%)
& Gap (\%) & Cross. & Hull Viol. (\%)
& Gap (\%) & Cross. & Hull Viol. (\%)
& Gap (\%) & Cross. & Hull Viol. (\%)
& Gap (\%) & Cross. & Hull Viol. (\%)
& Gap (\%) & Cross. & Hull Viol. (\%) \\
\midrule

Clustered
& 9.45 & 29.76 & 69.00
& 9.58 & 13.44 & 65.00
& 66.42 & 212.07 & 75.00
& 75.53 & 188.73 & 40.00
& 37.36 & 38.42 & \textbf{7.00}
& \textbf{2.76} & \textbf{2.77} & 48.00 \\

Expansion
& 9.38 & 14.97 & 22.00
& 5.58 & 5.99 & 13.00
& 56.92 & 92.10 & 7.00
& 30.35 & 8.11 & \textbf{0.00}
& 36.85 & 27.91 & \textbf{0.00}
& \textbf{2.37} & \textbf{1.94} & 6.00 \\

Explosion
& 4.65 & 4.78 & 41.00
& 3.83 & 3.60 & 11.00
& 52.55 & 73.73 & 4.00
& 31.19 & 8.31 & \textbf{1.00}
& 38.35 & 29.75 & 3.00
& \textbf{2.45} & \textbf{1.69} & 15.00 \\

Grid
& 4.50 & 7.54 & 52.00
& 3.56 & 4.11 & 39.00
& 39.61 & 86.19 & 22.00
& 33.42 & 8.62 & \textbf{2.00}
& 32.19 & 26.03 & 11.00
& \textbf{1.79} & \textbf{2.49} & 35.00 \\

Implosion
& 8.16 & 20.62 & 73.00
& 7.96 & 8.81 & 50.00
& 50.52 & 152.14 & 38.00
& 49.82 & 14.59 & \textbf{1.00}
& 36.47 & 41.76 & 14.00
& \textbf{1.85} & \textbf{1.98} & 45.00 \\

Mixed
& 5.25 & 13.33 & 45.00
& 4.49 & 7.16 & 49.00
& 55.75 & 188.51 & 71.00
& 35.33 & 11.76 & \textbf{1.00}
& 38.32 & 44.73 & 11.00
& \textbf{2.12} & \textbf{3.24} & 34.00 \\

Uniform
& 4.51 & 6.80 & 45.00
& 3.21 & 4.24 & 48.00
& 51.84 & 155.64 & 32.00
& 34.82 & 11.93 & \textbf{7.00}
& 36.95 & 38.24 & \textbf{7.00}
& \textbf{1.76} & \textbf{2.02} & 33.00 \\

\bottomrule
\end{tabular}%
}
\end{table*}

\clearpage
\subsection{CVRP Behavioral Results}\label{subsec:cvrp-behavioral}

We compare L2C-Insert, LEHD, POMO and AM across CVRP sizes from 20 to
1000 and seven test distributions. We report the optimality gap relative
to high-quality reference solutions obtained using a modern implementation
of Hybrid Genetic Search (HGS), together with the geometric metrics defined
in Section~\ref{subsec:tsp-behavioral}. For CVRP, edge crossings and
convex-hull order violations are evaluated separately within each vehicle
route and then averaged across the routes and problem instances. Lower
values are better for all metrics, and the best result in each setting,
including ties, is shown in bold.

Tables~\ref{tab:cvrp-20-three-model-comparison}--%
\ref{tab:cvrp-1000-three-model-comparison} show that LEHD exhibits stronger
scale generalization on the larger instances. It achieves the lowest
optimality gap across all seven distributions on CVRP-200 and in six of
the seven distributions on both CVRP-500 and CVRP-1000. In contrast,
L2C-Insert generally produces fewer within-route crossings and lower
convex-hull violation rates, indicating greater local geometric consistency.

\begin{table*}[ht!]
\centering
\scriptsize
\setlength{\tabcolsep}{3.0pt}
\renewcommand{\arraystretch}{1.08}
\caption{Solution quality and within-route geometric consistency under
greedy decoding on CVRP-20 across seven test distributions.}
\label{tab:cvrp-20-three-model-comparison}
\resizebox{\textwidth}{!}{%
\begin{tabular}{@{}l*{12}{c}@{}}
\toprule
\multirow{2}{*}{Distribution}
& \multicolumn{3}{c}{L2C-Insert}
& \multicolumn{3}{c}{LEHD}
& \multicolumn{3}{c}{POMO}
& \multicolumn{3}{c}{AM} \\
\cmidrule(lr){2-4}
\cmidrule(lr){5-7}
\cmidrule(lr){8-10}
\cmidrule(lr){11-13}
& Gap (\%) & Cross./Route & Hull Viol. (\%)
& Gap (\%) & Cross./Route & Hull Viol. (\%)
& Gap (\%) & Cross./Route & Hull Viol. (\%)
& Gap (\%) & Cross./Route & Hull Viol. (\%) \\
\midrule

Clustered
& 6.99 & \textbf{0.02} & \textbf{1.42}
& \textbf{5.02} & 0.04 & 3.47
& 71.63 & 0.27 & 16.87
& 14.27 & 0.24 & 18.84 \\

Expansion
& 6.65 & \textbf{0.02} & \textbf{1.68}
& \textbf{5.59} & 0.04 & 3.45
& 66.78 & 0.21 & 14.07
& 12.97 & 0.21 & 16.50 \\

Explosion
& 6.80 & \textbf{0.01} & \textbf{1.10}
& \textbf{5.76} & 0.04 & 3.11
& 66.77 & 0.18 & 12.36
& 12.72 & 0.17 & 14.15 \\

Grid
& 6.91 & \textbf{0.01} & \textbf{1.15}
& \textbf{5.76} & 0.03 & 2.83
& 65.97 & 0.15 & 10.92
& 12.71 & 0.17 & 13.97 \\

Implosion
& 6.98 & \textbf{0.01} & \textbf{1.16}
& \textbf{5.82} & 0.03 & 2.96
& 66.68 & 0.16 & 11.14
& 12.80 & 0.17 & 13.97 \\

Mixed
& 7.17 & \textbf{0.02} & \textbf{1.31}
& \textbf{5.35} & 0.04 & 2.98
& 60.44 & 0.18 & 11.94
& 13.54 & 0.20 & 15.22 \\

Uniform
& 6.87 & \textbf{0.01} & \textbf{1.18}
& \textbf{5.71} & 0.03 & 2.93
& 66.29 & 0.16 & 11.07
& 12.68 & 0.17 & 13.80 \\

\bottomrule
\end{tabular}%
}
\end{table*}

\begin{table*}[ht!]
\centering
\scriptsize
\setlength{\tabcolsep}{3.0pt}
\renewcommand{\arraystretch}{1.08}
\caption{Solution quality and within-route geometric consistency under
greedy decoding on CVRP-50 across seven test distributions.}
\label{tab:cvrp-50-three-model-comparison}
\resizebox{\textwidth}{!}{%
\begin{tabular}{@{}l*{12}{c}@{}}
\toprule
\multirow{2}{*}{Distribution}
& \multicolumn{3}{c}{L2C-Insert}
& \multicolumn{3}{c}{LEHD}
& \multicolumn{3}{c}{POMO}
& \multicolumn{3}{c}{AM} \\
\cmidrule(lr){2-4}
\cmidrule(lr){5-7}
\cmidrule(lr){8-10}
\cmidrule(lr){11-13}
& Gap (\%) & Cross./Route & Hull Viol. (\%)
& Gap (\%) & Cross./Route & Hull Viol. (\%)
& Gap (\%) & Cross./Route & Hull Viol. (\%)
& Gap (\%) & Cross./Route & Hull Viol. (\%) \\
\midrule

Clustered
& 5.82 & \textbf{0.02} & \textbf{1.27}
& \textbf{5.19} & 0.04 & 2.90
& 13.76 & 0.18 & 12.49
& 11.96 & 0.31 & 19.25 \\

Expansion
& 5.24 & \textbf{0.02} & \textbf{1.27}
& \textbf{4.73} & 0.04 & 2.99
& 11.89 & 0.16 & 11.13
& 9.83 & 0.25 & 16.55 \\

Explosion
& 5.30 & \textbf{0.01} & \textbf{0.80}
& \textbf{4.80} & 0.03 & 2.37
& 10.40 & 0.13 & 9.18
& 9.12 & 0.19 & 13.36 \\

Grid
& 5.29 & \textbf{0.01} & \textbf{0.68}
& \textbf{5.01} & 0.03 & 2.19
& 10.17 & 0.12 & 8.45
& 8.72 & 0.17 & 11.53 \\

Implosion
& 5.33 & \textbf{0.01} & \textbf{0.73}
& \textbf{5.00} & 0.03 & 2.33
& 10.26 & 0.12 & 8.37
& 8.74 & 0.17 & 12.06 \\

Mixed
& 5.50 & \textbf{0.01} & \textbf{0.93}
& \textbf{4.51} & 0.03 & 2.53
& 16.40 & 0.13 & 9.31
& 9.95 & 0.23 & 14.27 \\

Uniform
& 5.36 & \textbf{0.01} & \textbf{0.68}
& \textbf{5.01} & 0.03 & 2.18
& 10.22 & 0.11 & 8.26
& 8.74 & 0.17 & 11.85 \\

\bottomrule
\end{tabular}%
}
\end{table*}

\begin{table*}[ht!]
\centering
\scriptsize
\setlength{\tabcolsep}{3.0pt}
\renewcommand{\arraystretch}{1.08}
\caption{Solution quality and within-route geometric consistency under
greedy decoding on CVRP-100 across seven test distributions.}
\label{tab:cvrp-100-three-model-comparison}
\resizebox{\textwidth}{!}{%
\begin{tabular}{@{}l*{12}{c}@{}}
\toprule
\multirow{2}{*}{Distribution}
& \multicolumn{3}{c}{L2C-Insert}
& \multicolumn{3}{c}{LEHD}
& \multicolumn{3}{c}{POMO}
& \multicolumn{3}{c}{AM} \\
\cmidrule(lr){2-4}
\cmidrule(lr){5-7}
\cmidrule(lr){8-10}
\cmidrule(lr){11-13}
& Gap (\%) & Cross./Route & Hull Viol. (\%)
& Gap (\%) & Cross./Route & Hull Viol. (\%)
& Gap (\%) & Cross./Route & Hull Viol. (\%)
& Gap (\%) & Cross./Route & Hull Viol. (\%) \\
\midrule

Clustered
& \textbf{4.96} & \textbf{0.02} & \textbf{1.38}
& 5.05 & 0.05 & 3.48
& 5.49 & 0.12 & 8.30
& 12.91 & 0.45 & 23.67 \\

Expansion
& 4.38 & \textbf{0.02} & \textbf{1.17}
& 4.43 & 0.04 & 3.26
& \textbf{3.91} & 0.10 & 6.76
& 9.70 & 0.32 & 17.83 \\

Explosion
& 4.12 & \textbf{0.01} & \textbf{0.77}
& 4.28 & 0.03 & 2.46
& \textbf{3.37} & 0.07 & 4.61
& 8.63 & 0.23 & 13.05 \\

Grid
& 3.87 & \textbf{0.01} & \textbf{0.54}
& 4.16 & 0.03 & 1.92
& \textbf{2.98} & 0.06 & 3.59
& 7.65 & 0.17 & 9.98 \\

Implosion
& 3.91 & \textbf{0.01} & \textbf{0.60}
& 4.29 & 0.03 & 2.02
& \textbf{3.03} & 0.06 & 3.70
& 7.76 & 0.17 & 10.38 \\

Mixed
& 4.20 & \textbf{0.01} & \textbf{0.96}
& 4.09 & 0.04 & 2.62
& \textbf{3.72} & 0.08 & 4.78
& 8.95 & 0.29 & 14.38 \\

Uniform
& 3.88 & \textbf{0.01} & \textbf{0.56}
& 4.18 & 0.03 & 1.88
& \textbf{2.98} & 0.06 & 3.73
& 7.67 & 0.17 & 9.87 \\

\bottomrule
\end{tabular}%
}
\end{table*}

\begin{table*}[ht!]
\centering
\scriptsize
\setlength{\tabcolsep}{3.0pt}
\renewcommand{\arraystretch}{1.08}
\caption{Solution quality and within-route geometric consistency under
greedy decoding on CVRP-200 across seven test distributions.}
\label{tab:cvrp-200-three-model-comparison}
\resizebox{\textwidth}{!}{%
\begin{tabular}{@{}l*{12}{c}@{}}
\toprule
\multirow{2}{*}{Distribution}
& \multicolumn{3}{c}{L2C-Insert}
& \multicolumn{3}{c}{LEHD}
& \multicolumn{3}{c}{POMO}
& \multicolumn{3}{c}{AM} \\
\cmidrule(lr){2-4}
\cmidrule(lr){5-7}
\cmidrule(lr){8-10}
\cmidrule(lr){11-13}
& Gap (\%) & Cross./Route & Hull Viol. (\%)
& Gap (\%) & Cross./Route & Hull Viol. (\%)
& Gap (\%) & Cross./Route & Hull Viol. (\%)
& Gap (\%) & Cross./Route & Hull Viol. (\%) \\
\midrule

Clustered
& 5.81 & \textbf{0.05} & \textbf{2.52}
& \textbf{5.11} & 0.09 & 5.27
& 10.50 & 0.31 & 13.50
& 21.96 & 0.93 & 37.94 \\

Expansion
& 5.97 & \textbf{0.05} & \textbf{2.44}
& \textbf{4.34} & 0.07 & 4.81
& 9.62 & 0.33 & 12.47
& 13.77 & 0.58 & 24.98 \\

Explosion
& 5.29 & \textbf{0.03} & \textbf{1.59}
& \textbf{3.84} & 0.05 & 3.53
& 9.26 & 0.28 & 10.58
& 12.09 & 0.39 & 17.16 \\

Grid
& 4.89 & \textbf{0.03} & \textbf{1.42}
& \textbf{3.43} & 0.04 & 2.65
& 9.00 & 0.23 & 8.58
& 10.52 & 0.26 & 12.05 \\

Implosion
& 5.15 & \textbf{0.07} & \textbf{3.17}
& \textbf{3.78} & \textbf{0.07} & 4.18
& 9.17 & 0.32 & 11.66
& 13.41 & 0.41 & 15.71 \\

Mixed
& 5.07 & \textbf{0.03} & \textbf{1.79}
& \textbf{3.57} & 0.07 & 3.89
& 9.51 & 0.34 & 13.69
& 11.91 & 0.54 & 20.32 \\

Uniform
& 4.90 & \textbf{0.03} & \textbf{1.35}
& \textbf{3.47} & 0.04 & 2.73
& 9.08 & 0.24 & 8.91
& 10.66 & 0.27 & 12.11 \\

\bottomrule
\end{tabular}%
}
\end{table*}

\begin{table*}[ht!]
\centering
\scriptsize
\setlength{\tabcolsep}{3.0pt}
\renewcommand{\arraystretch}{1.08}
\caption{Solution quality and within-route geometric consistency under
greedy decoding on CVRP-500 across seven test distributions.}
\label{tab:cvrp-500-three-model-comparison}
\resizebox{\textwidth}{!}{%
\begin{tabular}{@{}l*{12}{c}@{}}
\toprule
\multirow{2}{*}{Distribution}
& \multicolumn{3}{c}{L2C-Insert}
& \multicolumn{3}{c}{LEHD}
& \multicolumn{3}{c}{POMO}
& \multicolumn{3}{c}{AM} \\
\cmidrule(lr){2-4}
\cmidrule(lr){5-7}
\cmidrule(lr){8-10}
\cmidrule(lr){11-13}
& Gap (\%) & Cross./Route & Hull Viol. (\%)
& Gap (\%) & Cross./Route & Hull Viol. (\%)
& Gap (\%) & Cross./Route & Hull Viol. (\%)
& Gap (\%) & Cross./Route & Hull Viol. (\%) \\
\midrule

Clustered
& \textbf{6.00} & \textbf{0.18} & \textbf{6.27}
& 6.72 & 0.34 & 14.97
& 59.38 & 3.67 & 49.81
& 74.56 & 2.07 & 48.33 \\

Expansion
& 5.05 & \textbf{0.15} & \textbf{5.72}
& \textbf{4.80} & 0.25 & 13.20
& 52.37 & 4.25 & 54.58
& 33.24 & 1.86 & 48.68 \\

Explosion
& 4.75 & \textbf{0.11} & \textbf{4.16}
& \textbf{4.08} & 0.19 & 10.38
& 49.54 & 3.32 & 51.04
& 28.88 & 1.44 & 40.05 \\

Grid
& 4.24 & \textbf{0.08} & \textbf{3.30}
& \textbf{3.36} & 0.14 & 8.63
& 39.47 & 2.52 & 47.73
& 21.74 & 0.96 & 30.62 \\

Implosion
& 4.26 & \textbf{0.18} & \textbf{5.84}
& \textbf{3.58} & 0.27 & 10.93
& 44.40 & 3.22 & 51.94
& 33.76 & 1.08 & 28.25 \\

Mixed
& 5.14 & \textbf{0.15} & \textbf{5.79}
& \textbf{4.25} & 0.28 & 12.01
& 61.65 & 3.38 & 51.09
& 26.51 & 1.58 & 38.69 \\

Uniform
& 4.18 & \textbf{0.08} & \textbf{3.32}
& \textbf{3.31} & 0.14 & 8.25
& 40.03 & 2.59 & 47.90
& 21.85 & 1.00 & 30.66 \\

\bottomrule
\end{tabular}%
}
\end{table*}

\begin{table*}[ht!]
\centering
\scriptsize
\setlength{\tabcolsep}{3.0pt}
\renewcommand{\arraystretch}{1.08}
\caption{Solution quality and within-route geometric consistency under
greedy decoding on CVRP-1000 across seven test distributions.}
\label{tab:cvrp-1000-three-model-comparison}
\resizebox{\textwidth}{!}{%
\begin{tabular}{@{}l*{12}{c}@{}}
\toprule
\multirow{2}{*}{Distribution}
& \multicolumn{3}{c}{L2C-Insert}
& \multicolumn{3}{c}{LEHD}
& \multicolumn{3}{c}{POMO}
& \multicolumn{3}{c}{AM} \\
\cmidrule(lr){2-4}
\cmidrule(lr){5-7}
\cmidrule(lr){8-10}
\cmidrule(lr){11-13}
& Gap (\%) & Cross./Route & Hull Viol. (\%)
& Gap (\%) & Cross./Route & Hull Viol. (\%)
& Gap (\%) & Cross./Route & Hull Viol. (\%)
& Gap (\%) & Cross./Route & Hull Viol. (\%) \\
\midrule

Clustered
& \textbf{10.77} & 1.04 & \textbf{21.95}
& 10.84 & \textbf{1.02} & 26.28
& 265.92 & 19.10 & 75.78
& 303.36 & 1.82 & 22.42 \\

Expansion
& 6.58 & \textbf{0.72} & \textbf{16.67}
& \textbf{6.08} & 0.82 & 28.04
& 308.93 & 19.57 & 73.11
& 118.61 & 2.64 & 37.37 \\

Explosion
& 6.83 & \textbf{0.49} & \textbf{13.38}
& \textbf{5.24} & 0.54 & 24.21
& 304.22 & 17.29 & 76.34
& 123.33 & 1.92 & 32.32 \\

Grid
& 5.78 & \textbf{0.30} & \textbf{9.32}
& \textbf{4.25} & 0.42 & 20.62
& 303.32 & 18.03 & 81.67
& 61.48 & 1.94 & 39.11 \\

Implosion
& 5.36 & \textbf{0.60} & \textbf{14.41}
& \textbf{4.21} & 0.77 & 23.57
& 289.05 & 18.20 & 80.75
& 109.97 & 1.52 & 26.42 \\

Mixed
& 8.58 & 1.14 & 22.14
& \textbf{6.25} & \textbf{0.86} & \textbf{21.93}
& 236.46 & 12.30 & 65.42
& 61.92 & 2.92 & 40.06 \\

Uniform
& 5.42 & \textbf{0.31} & \textbf{8.81}
& \textbf{4.02} & 0.40 & 20.81
& 301.74 & 17.49 & 80.67
& 63.87 & 2.01 & 38.81 \\

\bottomrule
\end{tabular}%
}
\end{table*}


\subsection{Sensitivity}
\label{subsec:sensitivity}

We evaluate the rotational robustness of L2C-Insert, LEHD, POMO,
and AM. For model $m$ and instance $i$, let $g_{i,\alpha}^{(m)}$
denote the optimality gap after rotating the instance by angle
$\alpha$. We consider
$\mathcal{A}=\{0^\circ,60^\circ,120^\circ,180^\circ,210^\circ,270^\circ\}$.
The rotation sensitivity of each instance is defined as
$s_i^{(m)}=\max_{\alpha\in\mathcal{A}}g_{i,\alpha}^{(m)}
-\min_{\alpha\in\mathcal{A}}g_{i,\alpha}^{(m)}$.
We report the mean sensitivity over the $M$ evaluated instances as
$S^{(m)}=\frac{1}{M}\sum_{i=1}^{M}s_i^{(m)}$.

Lower values indicate greater rotational robustness.
Tables~\ref{tab:rotation-sensitivity-small-tsp} and
\ref{tab:rotation-sensitivity-large-tsp} show that L2C-Insert
generally has the lowest sensitivity at smaller problem sizes,
while LEHD remains robust at larger scales. On TSP-500 and
TSP-1000, LEHD achieves the lowest sensitivity in five and four
of the seven distributions, respectively. POMO is generally more
sensitive than LEHD, while AM exhibits particularly high sensitivity
in several settings. The best result in each row, including ties,
is shown in bold.

\begin{table*}[ht!]
\centering

\caption{Rotation sensitivity across test distributions for
TSP-20, TSP-50, and TSP-100.}
\label{tab:rotation-sensitivity-small-tsp}
\resizebox{\textwidth}{!}{%
\begin{tabular}{@{}lcccc@{\hspace{8pt}}cccc@{\hspace{8pt}}cccc@{}}
\toprule
\multirow{2}{*}{Distribution}
& \multicolumn{4}{c}{TSP-20}
& \multicolumn{4}{c}{TSP-50}
& \multicolumn{4}{c}{TSP-100} \\
\cmidrule(lr){2-5}
\cmidrule(lr){6-9}
\cmidrule(lr){10-13}
& L2C-Insert & LEHD & POMO & AM
& L2C-Insert & LEHD & POMO & AM
& L2C-Insert & LEHD & POMO & AM \\
\midrule

Clustered
& \textbf{0.91} & 1.17 & 22.97 & 61.24
& \textbf{1.08} & 1.77 & 9.03 & 72.37
& \textbf{1.56} & 2.39 & 11.61 & 79.98 \\

Expansion
& 13.21 & \textbf{7.69} & 19.82 & 42.53
& \textbf{5.00} & 6.52 & 8.64 & 22.26
& \textbf{4.25} & 5.62 & 12.76 & 20.04 \\

Explosion
& 7.70 & \textbf{3.75} & 17.03 & 32.03
& \textbf{0.77} & 1.23 & 4.10 & 11.90
& \textbf{0.83} & 1.16 & 6.30 & 8.75 \\

Grid
& 6.58 & \textbf{2.81} & 16.40 & 26.12
& \textbf{0.82} & 1.11 & 2.44 & 6.90
& \textbf{0.81} & 1.03 & 1.76 & 4.81 \\

Implosion
& 6.21 & \textbf{2.84} & 16.51 & 27.93
& \textbf{0.84} & 1.12 & 2.55 & 7.15
& \textbf{0.78} & 1.09 & 1.80 & 4.89 \\

Mixed
& 6.79 & \textbf{4.53} & 21.64 & 34.33
& \textbf{1.68} & 2.00 & 11.78 & 11.65
& \textbf{1.52} & 1.82 & 18.09 & 6.44 \\

Uniform
& 6.79 & \textbf{2.97} & 16.21 & 27.41
& \textbf{0.86} & 1.12 & 2.46 & 7.16
& \textbf{0.80} & 1.03 & 1.71 & 4.84 \\

\bottomrule
\end{tabular}
}
\end{table*}

\begin{table*}[ht!]
\centering

\caption{Rotation sensitivity across test distributions for
TSP-200, TSP-500, and TSP-1000.}
\label{tab:rotation-sensitivity-large-tsp}
\resizebox{\textwidth}{!}{%
\begin{tabular}{@{}lcccc@{\hspace{8pt}}cccc@{\hspace{8pt}}cccc@{}}
\toprule
\multirow{2}{*}{Distribution}
& \multicolumn{4}{c}{TSP-200}
& \multicolumn{4}{c}{TSP-500}
& \multicolumn{4}{c}{TSP-1000} \\
\cmidrule(lr){2-5}
\cmidrule(lr){6-9}
\cmidrule(lr){10-13}
& L2C-Insert & LEHD & POMO & AM
& L2C-Insert & LEHD & POMO & AM
& L2C-Insert & LEHD & POMO & AM \\
\midrule

Clustered
& \textbf{1.94} & 2.38 & 16.87 & 96.41
& 3.03 & \textbf{2.93} & 20.59 & 101.45
& \textbf{3.45} & 3.46 & 24.60 & 113.51 \\

Expansion
& \textbf{4.82} & 7.18 & 5.66 & 7.46
& 24.23 & 11.62 & 15.75 & \textbf{9.25}
& 92.33 & 12.44 & 14.95 & \textbf{9.93} \\

Explosion
& \textbf{1.46} & 2.55 & 6.14 & 6.22
& \textbf{4.44} & 6.34 & 16.15 & 6.69
& 21.71 & 9.71 & 14.35 & \textbf{6.42} \\

Grid
& \textbf{1.05} & 1.10 & 6.00 & 4.22
& 1.89 & \textbf{1.56} & 10.30 & 4.21
& 2.65 & \textbf{1.94} & 9.28 & 3.72 \\

Implosion
& \textbf{1.67} & 1.87 & 6.68 & 174.96
& 2.67 & \textbf{2.36} & 12.24 & 193.60
& 3.30 & \textbf{2.64} & 10.32 & 269.56 \\

Mixed
& \textbf{1.53} & \textbf{1.53} & 19.15 & 5.06
& 2.55 & \textbf{2.05} & 23.34 & 5.28
& 2.99 & \textbf{2.30} & 28.46 & 4.82 \\

Uniform
& \textbf{1.18} & 1.20 & 5.65 & 5.04
& 1.92 & \textbf{1.76} & 12.57 & 5.30
& 2.89 & \textbf{2.46} & 10.36 & 4.99 \\

\bottomrule
\end{tabular}
}
\end{table*}
\clearpage
\section{Extended Mechanistic Interpretability Results}
\label{sec:mechinterp-extended}

\subsection{Controlling for Nearest-Neighbor Heuristics in LEHD}
\label{subsec:lehd_non_nn_probe}

To examine whether LEHD's multi-step future-action probe performance is
primarily driven by a simple nearest-neighbor heuristic, we restrict the
evaluation to decoding states where LEHD's own greedy next action is not
the nearest currently unvisited node. Results for the final two decoder
layers, L5 and L6, are reported in
Table~\ref{tab:lehd_future_probe_non_nn}.

Probe accuracy decreases under this restriction, but remains substantial
for multiple future steps. On TSP-500, for instance, L5 achieves 56.2\%
Top-1 accuracy at $h=2$ and 30.3\% at $h=3$, even though the model's
immediate action is explicitly non-nearest-neighbor. The same qualitative
pattern holds for TSP-50 and TSP-100. Thus, the future-action signal
accessible from LEHD's decoder representations is not reducible to a
simple nearest-neighbor rule. These results indicate that the
future-action information encoded in LEHD's decoder representations cannot
be explained solely by a simple nearest-neighbor heuristic and instead
reflects richer information about the model's subsequent trajectory.

\begin{table}[ht!]
\centering
\footnotesize
\caption{
Top-1 future-node probe accuracy (\%) for LEHD on TSP-50, TSP-100,
and TSP-500. \emph{Own-Trajectory Future} evaluates prediction of
LEHD's own future nodes from its greedy decoding trajectory.
\emph{Own-Trajectory Future (Non-NN)} restricts evaluation to states
where LEHD's greedy next action is not the nearest currently unvisited node.
}
\label{tab:lehd_future_probe_non_nn}

\resizebox{\linewidth}{!}{%
\begin{tabular}{llrrrrrrrrrrrr}
\toprule
&
&
\multicolumn{6}{c}{Own-Trajectory Future}
&
\multicolumn{6}{c}{Own-Trajectory Future (Non-NN)}
\\
\cmidrule(lr){3-8}
\cmidrule(lr){9-14}

Size
& Layer
& $h{=}1$
& $h{=}2$
& $h{=}3$
& $h{=}4$
& $h{=}5$
& $h{=}6$
& $h{=}1$
& $h{=}2$
& $h{=}3$
& $h{=}4$
& $h{=}5$
& $h{=}6$
\\
\midrule

TSP-50
& L5
& 95.6
& 86.5
& 60.5
& 32.4
& 22.5
& 17.2
& 85.4
& 70.5
& 40.2
& 30.6
& 19.6
& 16.0
\\

TSP-50
& L6
& 99.8
& 84.2
& 54.1
& 31.0
& 21.4
& 15.8
& 99.4
& 67.6
& 38.6
& 27.9
& 18.9
& 14.8
\\

\midrule

TSP-100
& L5
& 95.6
& 86.0
& 58.1
& 32.9
& 22.4
& 16.7
& 83.4
& 68.4
& 39.4
& 30.1
& 19.7
& 15.2
\\

TSP-100
& L6
& 99.9
& 82.4
& 50.4
& 31.2
& 20.9
& 15.5
& 99.8
& 65.1
& 35.5
& 27.7
& 19.0
& 14.3
\\

\midrule

TSP-500
& L5
& 94.5
& 77.7
& 43.6
& 28.2
& 20.4
& 15.4
& 78.2
& 56.2
& 30.3
& 22.5
& 17.4
& 13.7
\\

TSP-500
& L6
& 99.9
& 76.0
& 40.0
& 25.0
& 17.9
& 13.8
& 99.8
& 56.6
& 28.3
& 19.8
& 14.8
& 11.1
\\

\bottomrule
\end{tabular}%
}
\end{table}

\subsection{Behavioral-Causal Node Perturbation}\label{subsec:behavioral-causal}

In addition to the counterfactual steering experiment described in \Secref{subsec:counterfactual-steering}, we conduct a second causal experiment that does not rely on probes or manual modifications to the model's embeddings. Our setup compares the model's greedy rollout branch, \(a_1, a_2, \ldots\), with the greedy rollout starting from the second-highest-probability node, \(a'_1, a'_2, \ldots\). We perturb the coordinates of nodes \(a'_2, a'_3, \ldots, a'_5\), and then observe how the logit gap between \(a_1\) and \(a'_1\) changes at each step \(t\):
\[
\Delta G_t
=
\left[z_{\mathrm{steered}}(a'_1)-z_{\mathrm{steered}}(a_1)\right]
-
\left[z_{\mathrm{clean}}(a'_1)-z_{\mathrm{clean}}(a_1)\right],
\]
where \(z_{\mathrm{steered}}(\cdot)\) and \(z_{\mathrm{clean}}(\cdot)\) denote the logits under the perturbed and clean inputs, respectively.

We consider four perturbation categories:
\begin{itemize}
    \item The \textbf{toward} category shortens the distance between consecutive nodes \(a'_i\) and \(a'_{i-1}\) using an interpolation coefficient \(\alpha\):
    \[
    r_{a'_i} \leftarrow (1-\alpha) r_{a'_i} + \alpha r_{a'_{i-1}},
    \]
    where \(r_a\) denotes the location vector of node \(a\).
    \item Similarly, the \textbf{away} category increases the distance between consecutive nodes \(a'_i\) and \(a'_{i-1}\) using an extrapolation coefficient \(\alpha\).
    \item The \textbf{random direction} category perturbs the node in a random direction.
    \item The \textbf{random node} category perturbs a node in a far horizon, e.g., \(a'_{10}\).
\end{itemize}

\begin{figure*}[ht!]
    \centering

    \begin{subfigure}[t]{0.48\textwidth}
        \centering
        \includegraphics[
            width=\textwidth
        ]{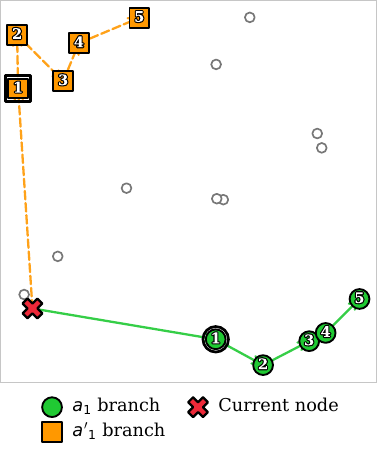}
        \caption{Before perturbation}
        \label{fig:causal-perturbation-sample-before}
    \end{subfigure}
    \begin{subfigure}[t]{0.48\textwidth}
        \centering
        \includegraphics[
            width=\textwidth
        ]{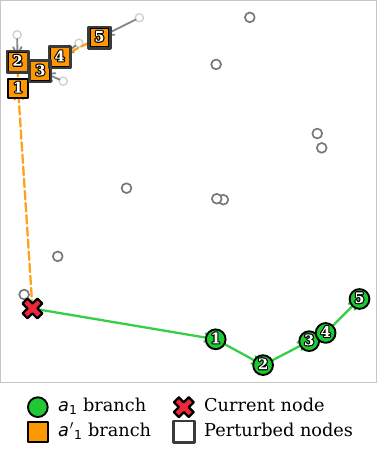}
        \caption{After perturbation}
        \label{fig:causal-perturbation-sample-after}
    \end{subfigure}

    \caption{A sample visualization of node perturbation. \textbf{(a)} Shows the original instance. \textbf{(b)} Shows the perturbed instance.}
    \label{fig:causal-perturbation-sample}
\end{figure*}

In most scenarios, there is only one plausible next node, with selection probability \(\approx 1\), and the second-best node is clearly poor. We therefore restrict our analysis to scenarios in which the second-most-probable node has a reasonable chance, defined as selection probability \(>0.20\). This makes the perturbations more meaningful. A sample visualization is presented in \Figref{fig:causal-perturbation-sample}

\begin{figure*}[ht!]
    \centering
    \includegraphics[
        width=\textwidth
    ]{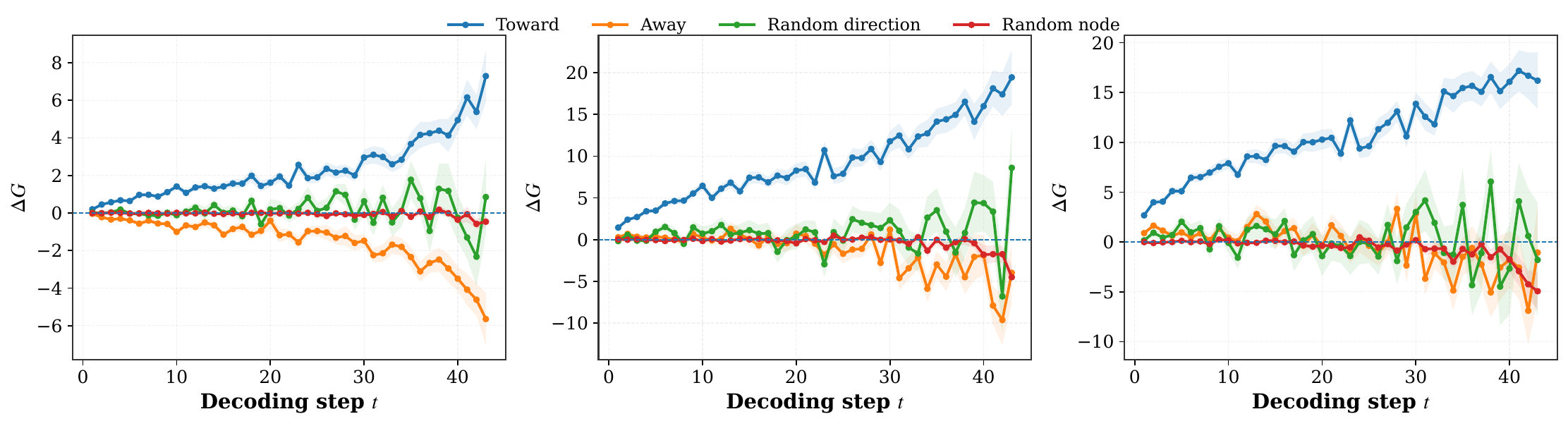}
    \caption{LEHD \(\Delta G\) with probability threshold.}
    \label{fig:lehd-deltag-prob}
\end{figure*}

In \Figref{fig:lehd-deltag-prob}, we observe that the gap between the most probable node, i.e., the original next node, and the second-highest-probability node in LEHD decreases in the \textbf{toward} category, where the distance between consecutive nodes along that horizon is shortened. This indicates that LEHD's preference shifts toward the second-most-probable branch. In contrast, we observe no meaningful change for the other perturbation categories, signaling that LEHD's decision remains relatively unchanged in those cases. Interestingly, this perturbation becomes even more effective as the model progresses through decoding.

\begin{figure*}[ht!]
    \centering
    \includegraphics[
        width=\textwidth
    ]{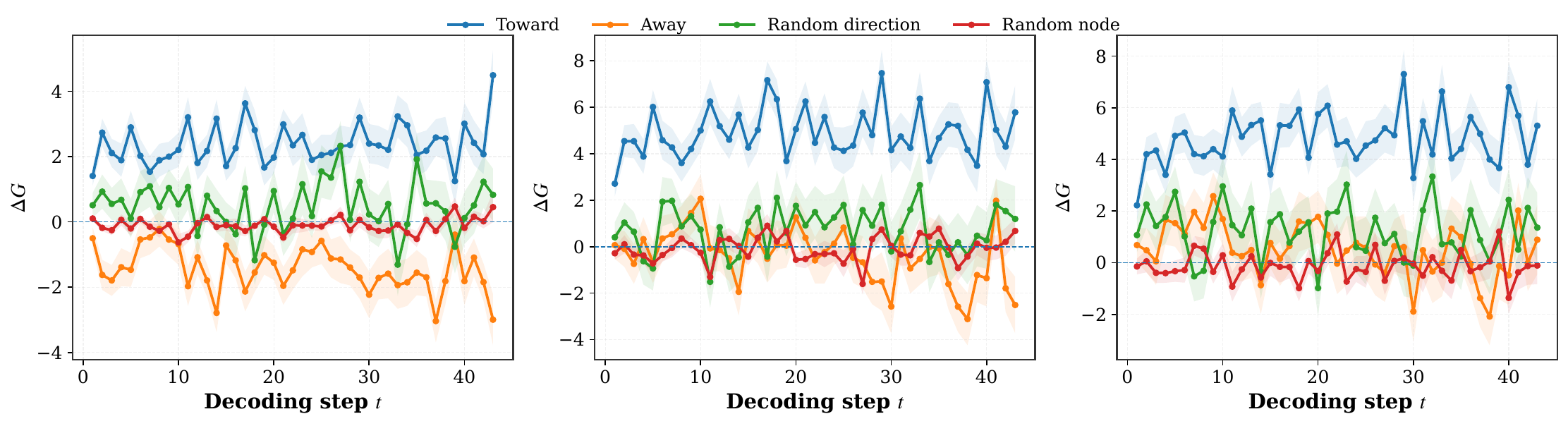}
    \caption{POMO \(\Delta G\) with probability threshold.}
    \label{fig:pomo-deltag-prob}
\end{figure*}

We also conducted the same experiment for POMO, with results shown in \Figref{fig:pomo-deltag-prob}. Although the \textbf{toward} category also produces a larger effect than the other categories, its value is much smaller than that of LEHD. This shows that LEHD is much more sensitive to the locations of nodes beyond the immediate next node than POMO is, highlighting the difference in look-ahead capabilities discussed in \Secref{sec:lehd}. Moreover, the effect of perturbation is relatively constant at every decoding step in POMO, unlike in LEHD, further highlighting POMO's lack of decoding-step sensitivity.

These experiments, together with the counterfactual steering experiment presented in \Secref{subsec:counterfactual-steering}, provide causal evidence for the look-ahead mechanism in LEHD.

\subsection{Assessing Off-Manifold Effects in Probe-Guided Steering}
\label{app:steering_off_manifold}

A potential concern with the probe-guided counterfactual steering
experiment in Section~\ref{subsec:counterfactual-steering} is that
the observed behavioral changes may result from pushing internal
representations away from the model's naturally occurring
representation manifold, rather than from selectively modifying
the future-action information under investigation. To assess
the extent of such off-manifold perturbations, we measure three
complementary representation-space diagnostics: the ratio of
steered to clean representation norms, the cosine similarity
between steered and clean representations, and the relative
norm of the perturbation. These quantities are computed for
the targeted future-node representations across the intervened
decoder layers and horizons. Table~\ref{tab:steering_representation_shift}
reports these diagnostics across steering strengths to assess
whether the interventions substantially alter the model's
natural representation geometry.

\begin{table}[ht!]
\centering
\caption{
Representation-space diagnostics for probe-guided
counterfactual steering on Uniform TSP-100 across
future horizons and decoder layers.
}
\label{tab:steering_representation_shift}
\begin{tabular}{cccc}
\toprule
$\alpha$
& Norm ratio
& Cosine similarity
& Relative perturbation \\
\midrule
3   & $1.031 \pm 0.002$ & $0.9873 \pm 0.0004$ & $0.151 \pm 0.002$ \\
\textbf{5}
    & $\mathbf{1.060 \pm 0.003}$
    & $\mathbf{0.9676 \pm 0.0008}$
    & $\mathbf{0.250 \pm 0.003}$ \\
10  & $1.156 \pm 0.006$ & $0.8959 \pm 0.0022$ & $0.495 \pm 0.007$ \\
\bottomrule
\end{tabular}
\end{table}

\subsection{Start-Node Navigation in AM}
\label{subsec:am-start-navigation}

Although all models are evaluated in a single-trajectory setting, AM differs from POMO and LEHD in how the starting node is selected. While POMO and LEHD begin from an arbitrary starting node, AM selects its starting node through its learned policy. At the initial decoding step, AM uses two learnable placeholder embeddings for the first and current nodes, enabling start selection to be learned jointly with subsequent route construction. AM and POMO also differ in their decoder conditioning: AM concatenates the first- and current-node embeddings and incorporates a separate graph-level representation, whereas POMO combines their projected embeddings through addition. These differences in start selection and learned decoder conditioning may contribute to AM's greater sensitivity to start-node interventions. 

\begin{figure*}[ht!]
    \centering
    \includegraphics[width=0.95\textwidth]
    {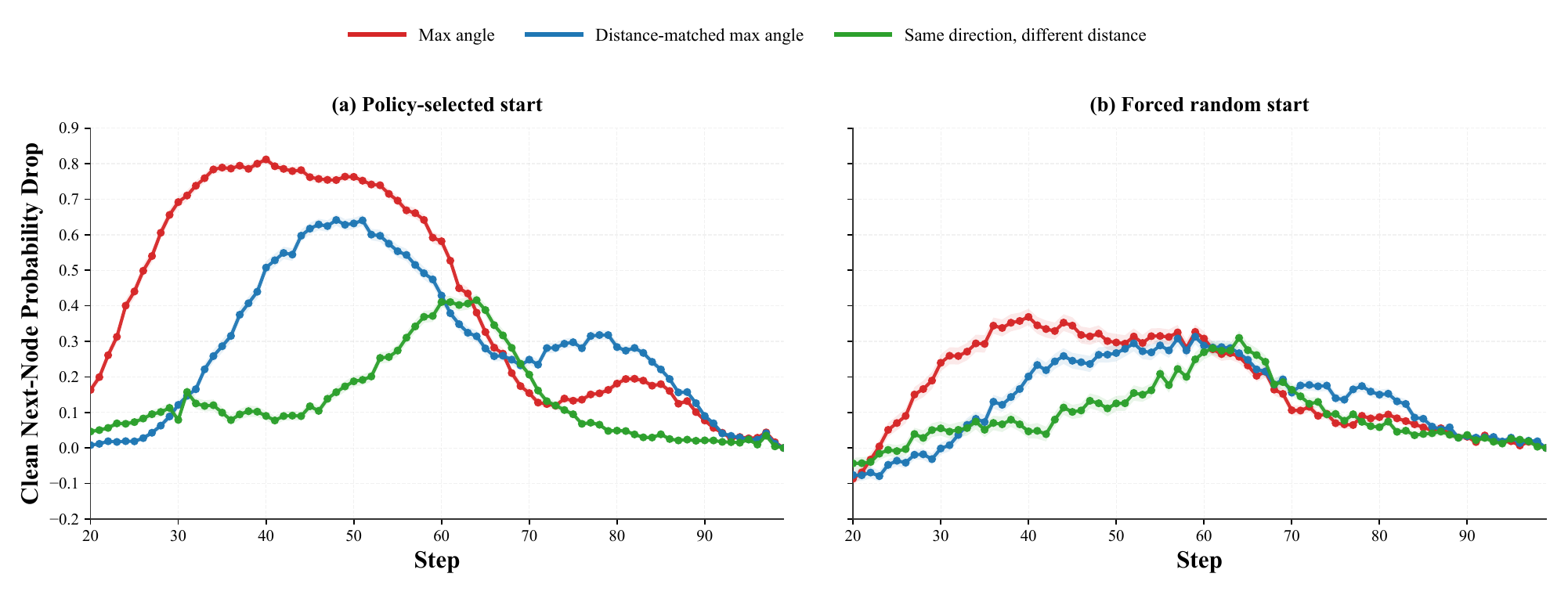}
    \caption{
    \textbf{Start-node intervention sensitivity in AM on Uniform TSP-100.}
    (a) Policy-selected start; (b) forced start at node~0,
    followed by greedy decoding.
    Previously visited donors are selected by maximum angular
    separation (red), maximum angle with matched distance (blue),
    or similar direction with different distance (green).
    The y-axis shows the mean drop in clean next-action probability
    after patching.
    }
    \label{fig:am-start-navigation-policy-vs-forced}
\end{figure*}

As shown in Figure~\ref{fig:am-start-navigation-policy-vs-forced},
start-node patching causes substantially smaller probability drops
when node~0 is prescribed as the starting node rather than selected
by AM's policy. This suggests that learned start selection may make
the start-node representation more influential in subsequent routing
decisions, although the two settings also produce different tour
trajectories. One possible explanation for the contrasting trends
in AM and LEHD is that AM relies more on start-node information
during early and middle tour construction, whereas LEHD may use it
more heavily as the tour approaches completion.

Under \textit{maximum-angle donor selection without distance constraints},
policy-selected starts yield lower LCS and higher Rev.\ LCS
than forced starts in Phase I
(Table~\ref{tab:am-start-lcs-phase1}), but this pattern is less
consistent in Phase II
(Table~\ref{tab:am-start-lcs-phase2}). Unlike LEHD
(Table~\ref{tab:start-patching-lcs-n100}), AM retains higher
forward- than reverse-order similarity even at large angles.
This may suggest a policy-dependent route-anchoring role in
AM, compared with a possible global-navigation reference
in LEHD.

\begin{table}[ht!]
\centering
\small

\begin{minipage}[t]{0.48\textwidth}
    \centering
    \small
    \setlength{\tabcolsep}{2.5pt}
    \renewcommand{\arraystretch}{1.08}

    \captionof{table}{
    AM rollout-order similarity on Uniform TSP-100 under
    maximum-angle donor selection.
    Phase I: steps 20--45.
    }
    \label{tab:am-start-lcs-phase1}

    \begin{tabular}{lrr|rr}
    \toprule
    \multirow{2}{*}{Angle Range}
    & \multicolumn{2}{c|}{Policy Start}
    & \multicolumn{2}{c}{Forced Start} \\
    \cmidrule(lr){2-3}\cmidrule(lr){4-5}
    & LCS
    & Rev.\ LCS
    & LCS
    & Rev.\ LCS \\
    \midrule

    $0^\circ$--$30^\circ$
    & \textbf{0.626} & 0.159
    & \textbf{0.680} & 0.137 \\

    $30^\circ$--$60^\circ$
    & \textbf{0.564} & 0.183
    & \textbf{0.583} & 0.172 \\

    $60^\circ$--$90^\circ$
    & \textbf{0.464} & 0.215
    & \textbf{0.522} & 0.191 \\

    $90^\circ$--$120^\circ$
    & \textbf{0.369} & 0.250
    & \textbf{0.484} & 0.204 \\

    $120^\circ$--$150^\circ$
    & \textbf{0.351} & 0.263
    & \textbf{0.435} & 0.221 \\

    $150^\circ$--$180^\circ$
    & \textbf{0.380} & 0.261
    & \textbf{0.437} & 0.227 \\

    \bottomrule
    \end{tabular}
\end{minipage}
\hfill
\begin{minipage}[t]{0.48\textwidth}
    \centering
    \small
    \setlength{\tabcolsep}{2.5pt}
    \renewcommand{\arraystretch}{1.08}

    \captionof{table}{
    AM rollout-order similarity on Uniform TSP-100 under
    maximum-angle donor selection.
    Phase II: steps 46--99.
    }
    \label{tab:am-start-lcs-phase2}

    \begin{tabular}{lrr|rr}
    \toprule
    \multirow{2}{*}{Angle Range}
    & \multicolumn{2}{c|}{Policy Start}
    & \multicolumn{2}{c}{Forced Start} \\
    \cmidrule(lr){2-3}\cmidrule(lr){4-5}
    & LCS
    & Rev.\ LCS
    & LCS
    & Rev.\ LCS \\
    \midrule

    $0^\circ$--$30^\circ$
    & -- & --
    & -- & -- \\

    $30^\circ$--$60^\circ$
    & \textbf{0.782} & 0.138
    & \textbf{0.748} & 0.145 \\

    $60^\circ$--$90^\circ$
    & \textbf{0.726} & 0.159
    & \textbf{0.711} & 0.163 \\

    $90^\circ$--$120^\circ$
    & \textbf{0.680} & 0.180
    & \textbf{0.689} & 0.175 \\

    $120^\circ$--$150^\circ$
    & \textbf{0.708} & 0.176
    & \textbf{0.688} & 0.178 \\

    $150^\circ$--$180^\circ$
    & \textbf{0.760} & 0.172
    & \textbf{0.685} & 0.182 \\

    \bottomrule
    \end{tabular}
\end{minipage}

\end{table}
\clearpage
\section{Visualization}\label{sec:visualization}
We present additional visualizations of tour-construction patterns,
routing behaviors, and internal representations of neural solvers.

\subsection{Latent-Space Organization}
\label{subsec:latent-space-node-role-alignment}

\begin{figure*}[ht!]
    \centering
    \includegraphics[
        width=\textwidth
    ]{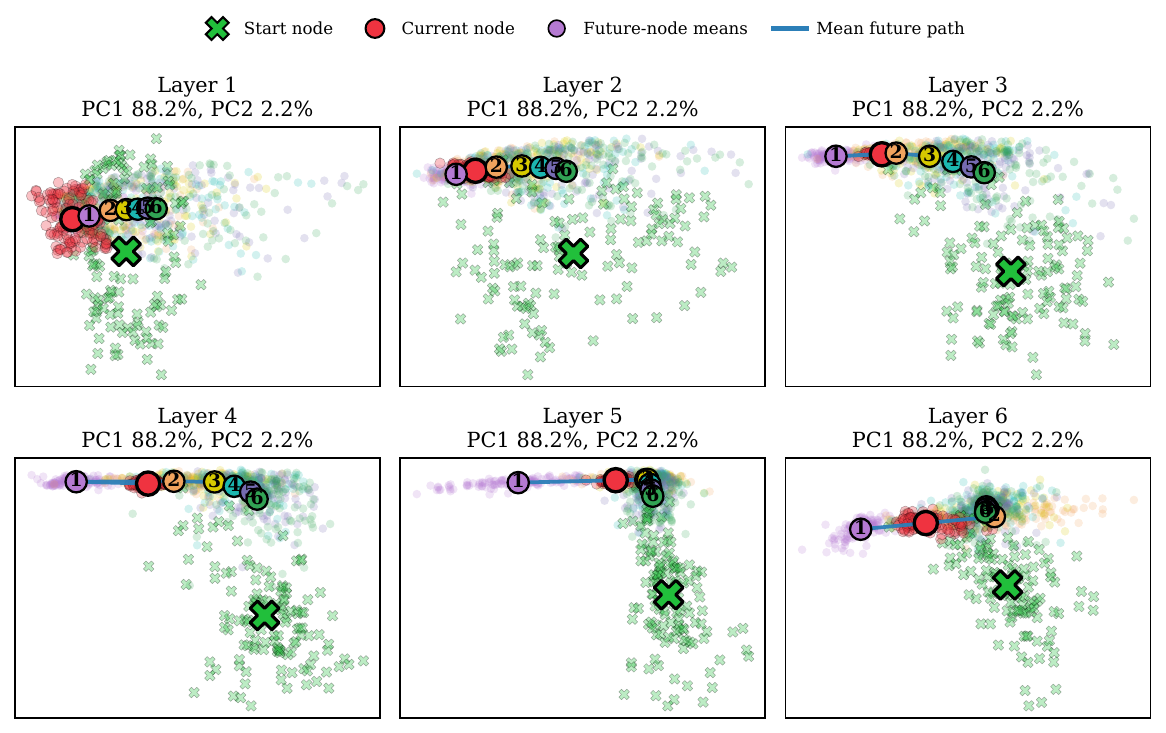}
    \caption{\textbf{Layer-wise PCA of LEHD node-role representations.}
    Faint points show instance-level representations across 128 Uniform
    TSP-50 instances, while larger markers indicate their means. Numbered
    markers denote LEHD's next six actions under greedy decoding.}
    \label{fig:lehd-latent-space-pca}
\end{figure*}

Figure~\ref{fig:lehd-latent-space-pca} provides a qualitative view of how
LEHD organizes node-role representations across decoder layers. A single
PCA basis is jointly fitted across all decoder layers, allowing direct
comparison of their representations in a shared latent space. The reported
explained-variance ratios correspond to this joint PCA. The faint markers
show instance-level representations, while the larger markers denote their
corresponding means. From Layer~2 onward, current-node representations
from different instances become concentrated in a compact shared region,
consistent with the cross-instance node alignment reported in the main text.
In contrast, the start-node representations remain more dispersed in the
earlier layers, but their distinct region becomes increasingly apparent from
Layer~4 onward, suggesting that the decoder progressively separates the
start node as a dedicated global reference. The mean future-node
representations also exhibit an approximately ordered trajectory, particularly
in the intermediate layers.

\subsection{Geometric Tour-Construction Patterns}
\label{subsec:geometric-tour-construction-patterns}

\begin{figure*}[ht!]
    \centering
    \includegraphics[
        width=\textwidth
    ]{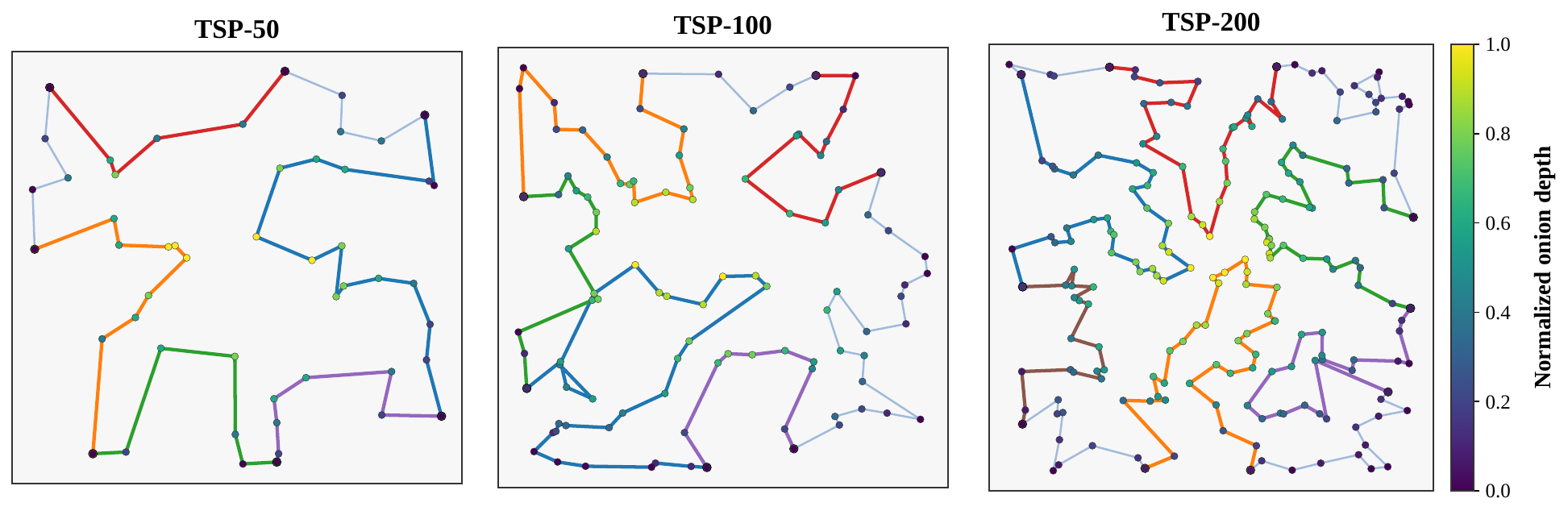}
    \caption{\textbf{Onion-depth structure of POMO tours.}
    Representative Uniform TSP-50, TSP-100, and TSP-200 rollouts illustrate
    POMO's repeated oscillation between outer and deeper onion layers. Colored
    segments highlight example outer-to-deep-to-outer excursions, while node
    colors indicate normalized onion depth.}
    \label{fig:pomo-onion-depth-triptych}
\end{figure*}

\begin{figure*}[ht!]
    \centering
    \includegraphics[
        width=\textwidth
    ]{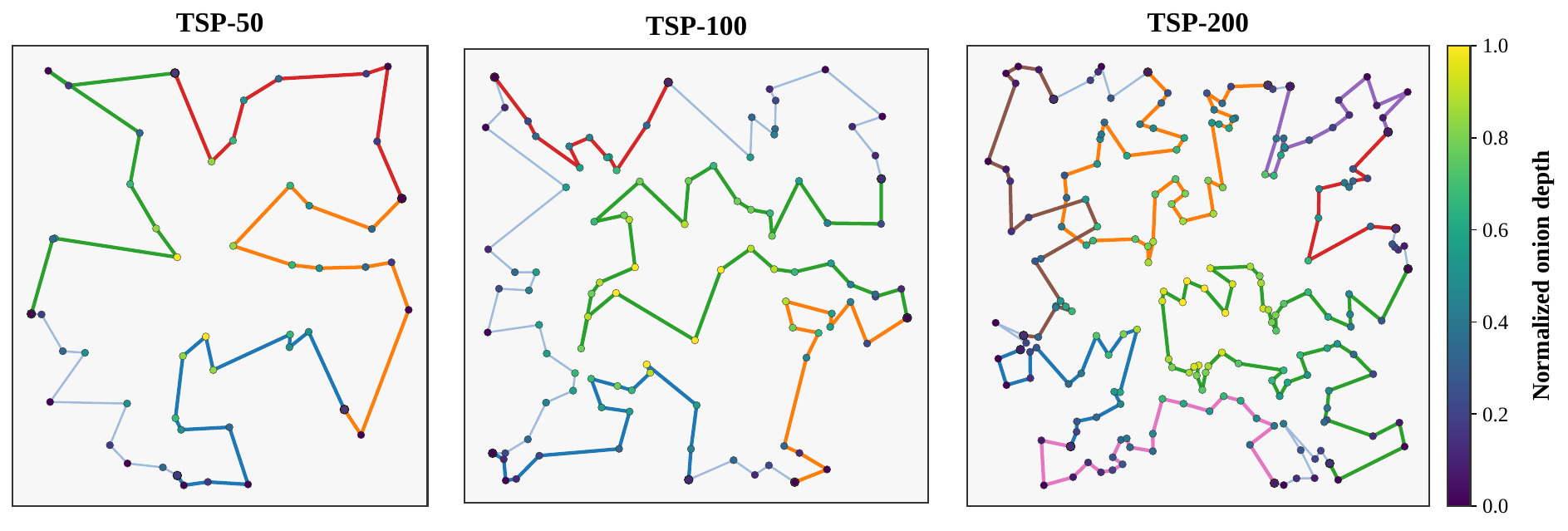}
    \caption{\textbf{Onion-depth structure of AM tours.}
    Selected Uniform TSP-50, TSP-100, and TSP-200 rollouts illustrate
    outer-to-deep-to-outer excursions in AM. Colored segments highlight
    example excursions, while node colors indicate normalized onion depth.}
    \label{fig:am-onion-depth-triptych}
\end{figure*}

Figures~\ref{fig:pomo-onion-depth-triptych} and
\ref{fig:am-onion-depth-triptych} provide representative examples of
POMO and AM rollouts across increasing problem sizes. Node colors indicate
normalized onion depth, with outer nodes assigned lower values and interior
nodes higher values. In both models, the highlighted segments reveal
outer-to-deep-to-outer excursions during tour construction. In POMO, this
pattern appears clearly already on TSP-50 and persists, becoming more
pronounced as the problem size increases to TSP-100 and TSP-200, consistent
with the oscillatory behavior quantified by the onion-depth analysis. The
corresponding AM examples show that similar excursions can also arise under
the same visualization, providing a qualitative comparison of geometric
tour-construction patterns across the two models, without implying that the
frequency or strength of these excursions is the same in AM and POMO.

\begin{figure*}[ht!]
    \centering

    \begin{subfigure}[t]{0.98\textwidth}
        \centering
        \includegraphics[
            width=\textwidth
        ]{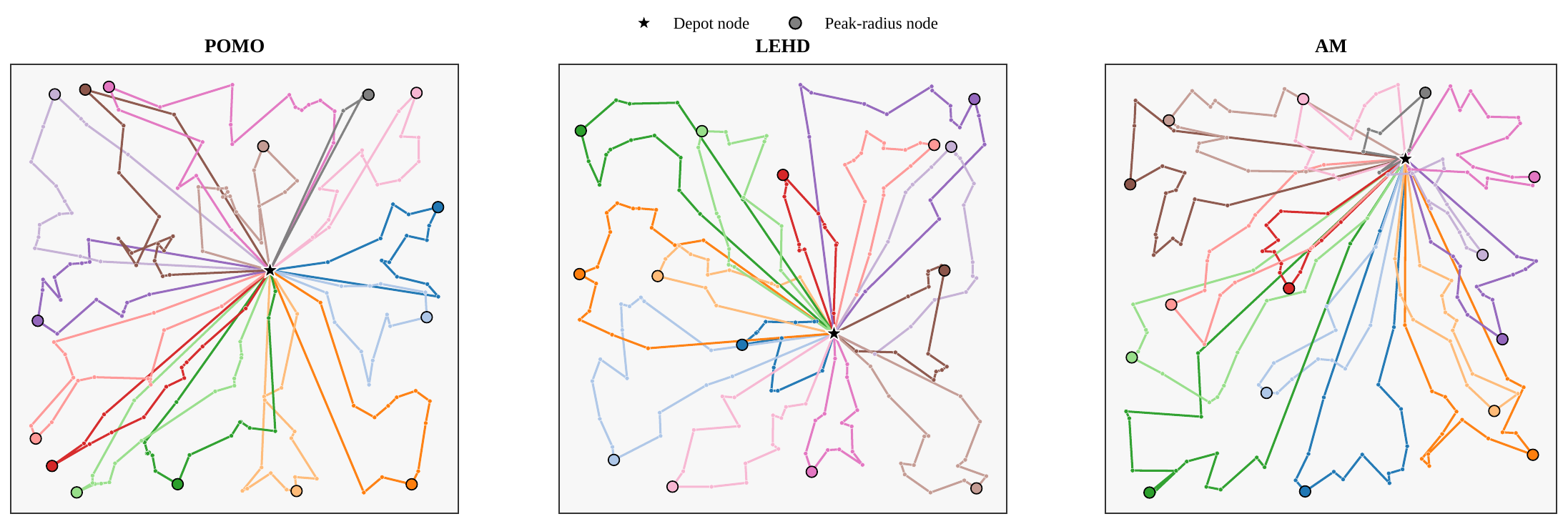}
        \caption{Uniform CVRP-200.}
        \label{fig:pomo-lehd-cvrp200-radial}
    \end{subfigure}

    \vspace{0.5em}

    \begin{subfigure}[t]{0.98\textwidth}
        \centering
        \includegraphics[
            width=\textwidth
        ]{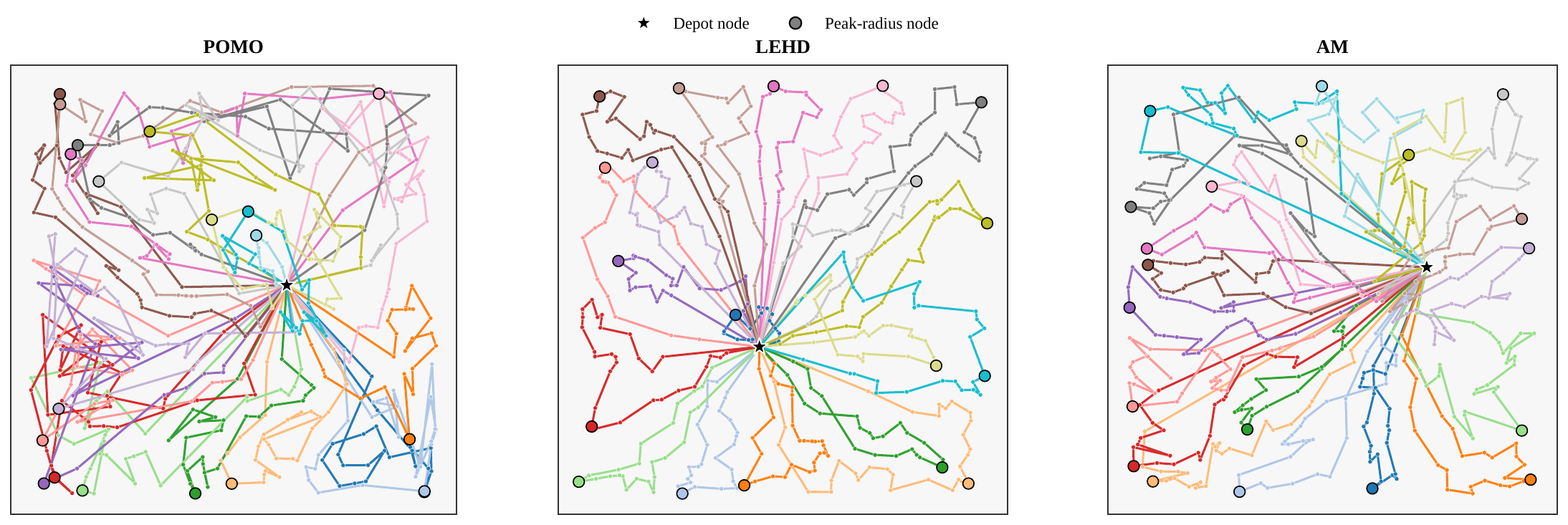}
        \caption{Uniform CVRP-500.}
        \label{fig:pomo-lehd-cvrp500-radial}
    \end{subfigure}

    \caption{\textbf{Radial route structure of POMO, LEHD, and AM on CVRP.}
    Selected Uniform CVRP-200 and CVRP-500 solutions are shown, with
    POMO on the left, LEHD in the center, and AM on the right.
    Each vehicle route is displayed in a distinct color; black stars
    indicate depot locations, and enlarged circular markers identify
    the farthest customer from the depot along each route.
    In these selected examples, radial backtracking is most visually
    pronounced in POMO, followed by AM, while LEHD exhibits more
    regular outward-then-inward route structures.}
    \label{fig:pomo-lehd-cvrp-radial-comparison}
\end{figure*}

Figure~\ref{fig:pomo-lehd-cvrp-radial-comparison} qualitatively compares
the radial organization of POMO, LEHD, and AM routes on Uniform CVRP-200
and CVRP-500. Among the selected high-backtracking examples, POMO exhibits
the most pronounced radial reversals, producing intertwined trajectories
that repeatedly move toward and away from the depot. AM shows an
intermediate pattern, with several routes displaying radial reversals
alongside more structured outward and return segments. In contrast, LEHD
exhibits the most regular radial organization in these examples, with
routes generally extending toward their peak-radius nodes before
returning to the depot. The differences become particularly visible on
CVRP-500, where the longer routes make the contrasting geometric
structures easier to observe. These visualizations illustrate the
differences in radial backtracking examined in the quantitative analysis.

\subsection{Attention Patterns During Decoding}
\label{subsec:attention-patterns-decoding}

\begin{figure}[ht!]
    \centering
    \includegraphics[width=\linewidth]{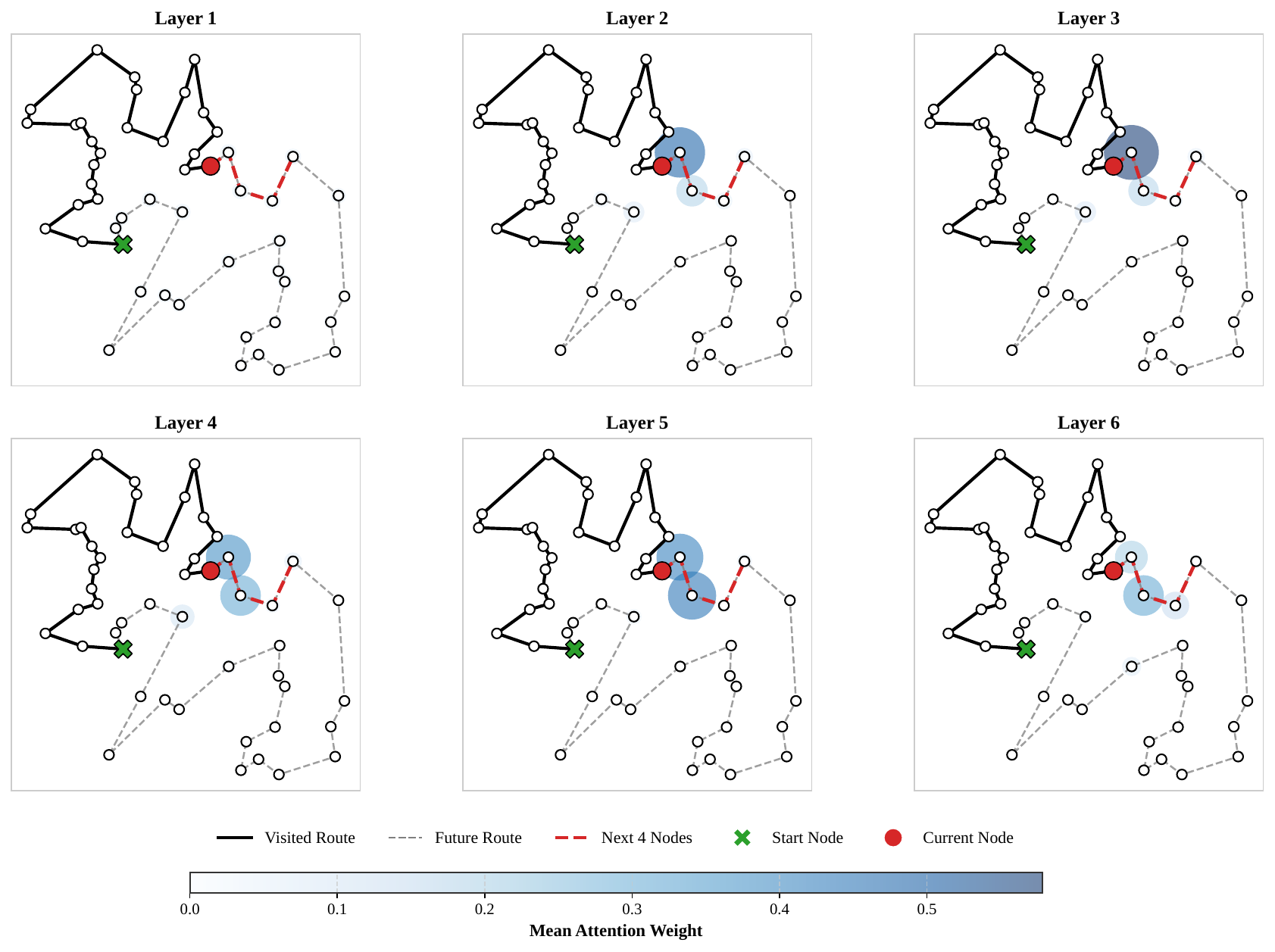}
    \caption{\textbf{Layer-wise attention heatmaps of the current node during decoding.} Each panel illustrates the attention weight distribution assigned by the current node (query) to remaining unvisited nodes (keys) for each decoder layer at a representative decoding step. Red dashed lines denote the nodes actually visited in subsequent steps.}
    \label{fig:lehd_attention_heatmap_layered}
\end{figure}

In \Figref{fig:lehd-attention-heatmap} of the main text, we visualized the attention heatmap for a single layer of LEHD to demonstrate that the model attends more strongly to nodes scheduled for future visits than to other unvisited nodes. In \Figref{fig:lehd_attention_heatmap_layered}, we provide a detailed layer-wise breakdown of this behavior across all decoder layers. The upcoming nodes in the ground-truth sequence are highlighted with red dashed lines. For visual clarity, self-attention from the current node to itself has been omitted. Notably, the attention distribution is more diffuse in the initial layer and becomes increasingly focused on upcoming nodes in subsequent layers, suggesting that future-action planning primarily occurs in the deeper decoding layers. Additionally, an animated visualization of the step-wise attention dynamics across layers is provided in \href{https://github.com/NCO-Interpretability/NCO-Interpretability}{https://github.com/NCO-Interpretability/NCO-Interpretability}.

\clearpage
\section{CVRP Results}\label{sec:cvrp-results}
In this section, we extend the interpretability experiments presented in Section~\ref{sec:experiments} to the Capacitated Vehicle Routing Problem (CVRP). Unlike the TSP, where a solution consists of a single continuous tour, CVRP solutions comprise multiple distinct routes that each originate and terminate at a central depot. Here, we evaluate whether the mechanistic and geometric properties observed in TSP generalize to individual routes within CVRP solutions.

\subsection{Geometric Trajectory Analysis}

\begin{figure*}[ht!]
    \centering
    \begin{subfigure}[t]{0.33\textwidth}
        \centering
        \includegraphics[width=\linewidth]{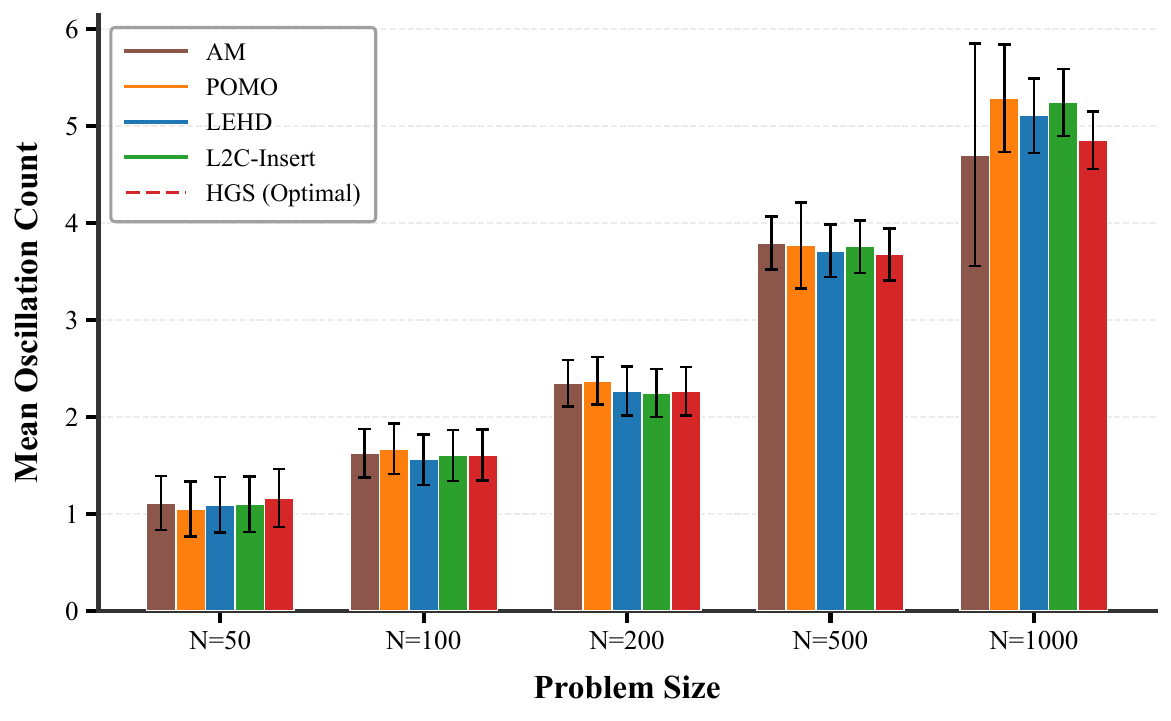}
        \caption{Mean onion-layer oscillations}
        \label{fig:cvrp_mean_oscillation}
    \end{subfigure}
    \begin{subfigure}[t]{0.33\textwidth}
        \centering
        \includegraphics[width=\linewidth]{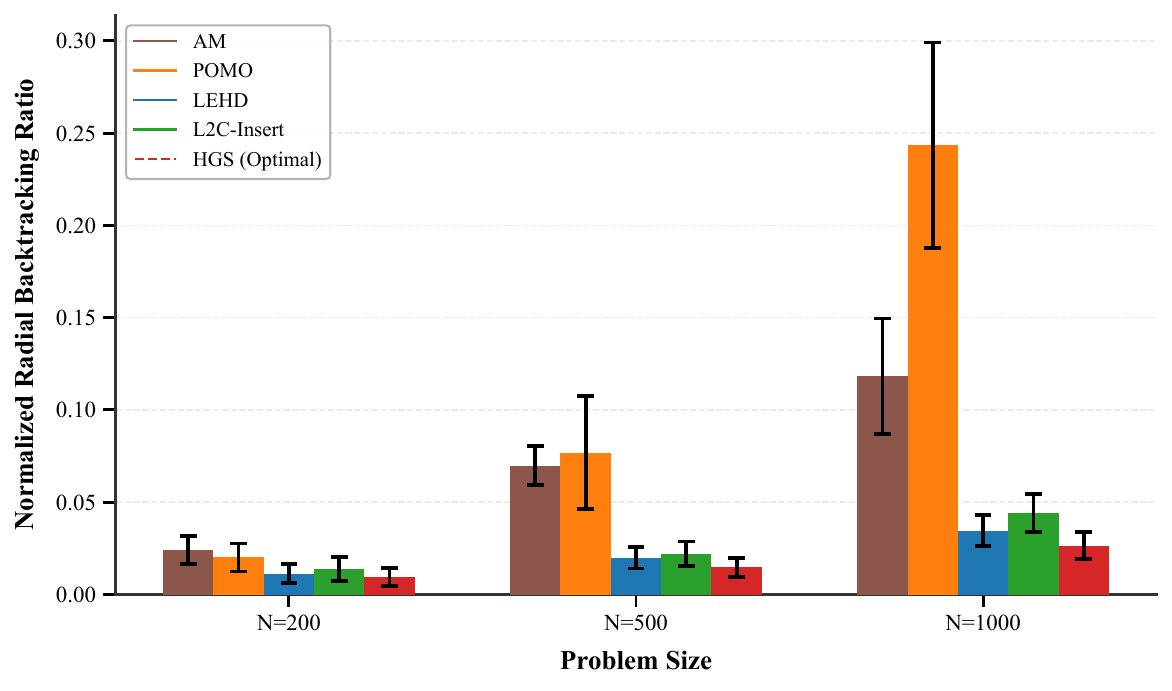}
        \caption{Radial route distortion}
        \label{fig:cvrp_radial_oscillation}
    \end{subfigure}
    \begin{subfigure}[t]{0.33\textwidth}
        \centering
        \includegraphics[width=\linewidth]{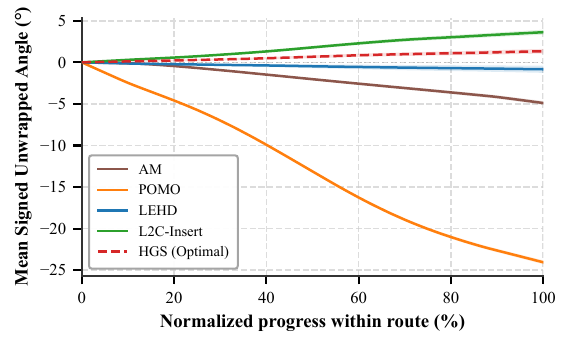}
        \caption{Angular orientation progression}
        \label{fig:cvrp_angle_progression}
    \end{subfigure}
    \caption{\textbf{Geometric trajectory properties on CVRP routes.}
    \textbf{(a)} Average onion-layer oscillations across problem scales; values remain low due to short individual route lengths.
    \textbf{(b)} Step-wise radial distortion along individual routes, showing heightened back-and-forth movement in POMO compared to baselines, with AM exhibiting intermediate distortion at larger scales.
    \textbf{(c)} Mean current-to-depot angle relative to each route's first edge, averaged within instances and then across instances, showing POMO's persistent clockwise progression and a weaker clockwise tendency in AM.}
    \label{fig:cvrp_polar_trajectory}
\end{figure*}

For the TSP, we established that POMO constructs tours via an oscillating clockwise progression. However, because individual routes in CVRP are substantially shorter than full TSP tours, even on large problem sizes, the average number of deep onion-layer oscillations per route is naturally low across all methods, rendering this metric less effective for distinguishing geometric biases (\Figref{fig:cvrp_mean_oscillation}).

To capture radial geometric distortion within shorter individual routes, we first identify the node furthest from the depot within a given route and use it as an anchor to bisect the route into two segments. For the first segment, we compute the depot distance of each node and count how often the distance \textit{decreases} between consecutive steps. For the second segment, we count how often the distance \textit{increases}. Intuitively, this measures the degree of radial back-and-forth distortion along the route. As shown in \Figref{fig:cvrp_radial_oscillation} and visually illustrated in \Figref{fig:pomo-lehd-cvrp-radial-comparison}, POMO exhibits higher radial distortion than LEHD, L2C-Insert, and HGS, while AM shows comparable distortion at smaller scales but substantially less at N=1000. Furthermore, POMO maintains a consistent clockwise progression during route construction, mirroring its TSP behavior (\Figref{fig:cvrp_angle_progression}). AM also exhibits a milder clockwise tendency on CVRP, in contrast to its counterclockwise progression on TSP.

\subsection{Future-Action Planning Probes}

\begin{figure*}[ht!]
    \centering
    \begin{subfigure}[t]{0.54\textwidth}
        \centering
        \includegraphics[width=\linewidth]{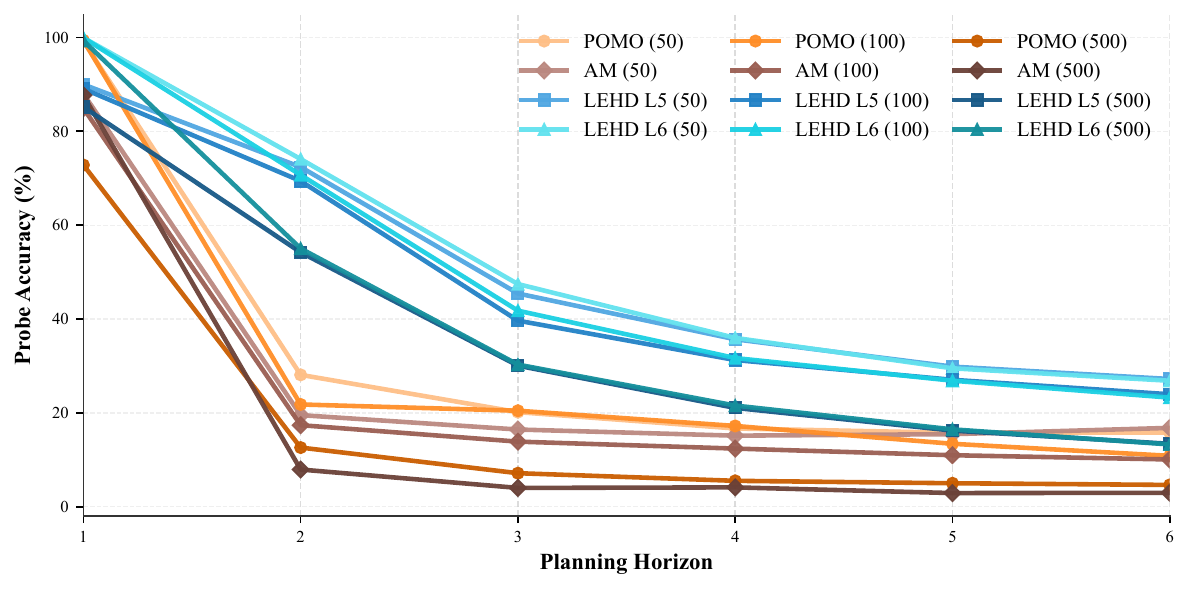}
        \caption{Future-action linear probing accuracy}
        \label{fig:cvrp_horizon_probe}
    \end{subfigure}
    \hfill
    \begin{subfigure}[t]{0.44\textwidth}
        \centering
        \includegraphics[width=\linewidth]{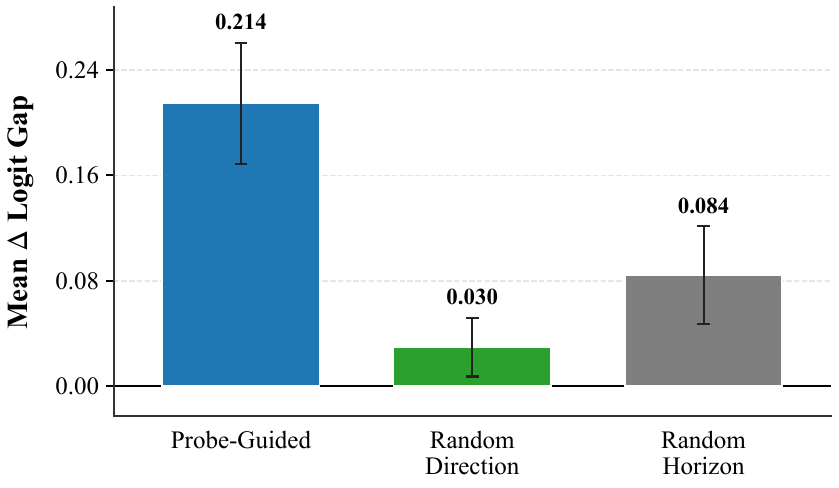}
        \caption{Probe-guided steering}
        \label{fig:cvrp_future_steering}
    \end{subfigure}

    \caption{\textbf{Future-action representations and causal steering on CVRP.}
    \textbf{(a)} Linear probing accuracy across planning horizons for LEHD, POMO, and AM, revealing differences in the accessibility of future-action information from decoder representations.
    \textbf{(b)} Change in the logit gap under probe-guided steering and control interventions on CVRP-100, measuring the effect of modifying future-node representations on the current decision.}
    \label{fig:cvrp_pomo_lehd_horizon_probe}
\end{figure*}

In Section~\ref{sec:lehd}, we showed that LEHD's internal
representations encode multi-step future node sequences for the TSP.
To determine whether this look-ahead capability extends to CVRP,
we trained linear probes to predict upcoming nodes within the
active route. For CVRP, the probe ranks all unvisited customers,
regardless of their immediate feasibility under the remaining
vehicle capacity. We exclude horizons that cross a depot return,
ensuring that each evaluated target belongs to the same route as
the current customer. As shown in \Figref{fig:cvrp_horizon_probe},
LEHD demonstrates strong predictive accuracy over short future
horizons, closely replicating the planning characteristics
observed on TSP. In contrast, both POMO and AM retain only limited
horizon-predictive information, with probe accuracy dropping
sharply beyond one-step prediction.

To investigate whether these future-action representations also influence the current decision, we extend our probe-guided steering experiment to CVRP. At each selected state, we identify the model's preferred next customer $H_1$ and its second-best alternative $H_1'$, then counterfactually force $H_1'$ and collect subsequent customers $H_2',\ldots,H_5'$ within the same vehicle route. We intervene on the representations of these future customers without directly modifying $H_1$ or $H_1'$. As shown in \Figref{fig:cvrp_future_steering}, steering LEHD's future-node representations increases the logit gap in favor of $H_1'$, with a larger mean effect than either the random-direction or disjoint-horizon control. These findings provide evidence that LEHD's accessible future-action information is causally connected to its current routing decisions in CVRP.

\subsection{Current and Start Node Contribution}

\begin{figure}[ht!]
    \centering
    \begin{subfigure}[t]{0.37\textwidth}
        \centering
        \includegraphics[width=\linewidth]{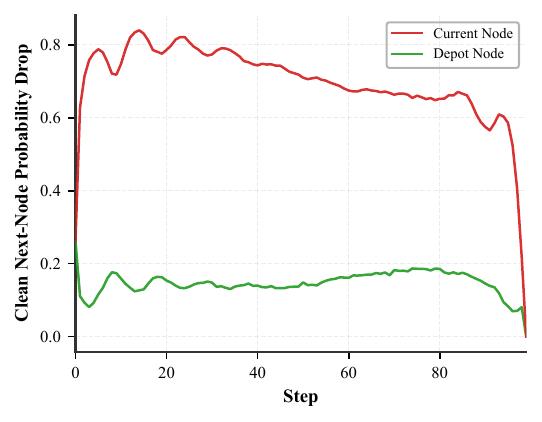}
        \caption{LEHD mean ablation impact}
        \label{fig:cvrp_mean_ablation}
    \end{subfigure}
    \begin{subfigure}[t]{0.61\textwidth}
        \centering
        \includegraphics[width=\linewidth]{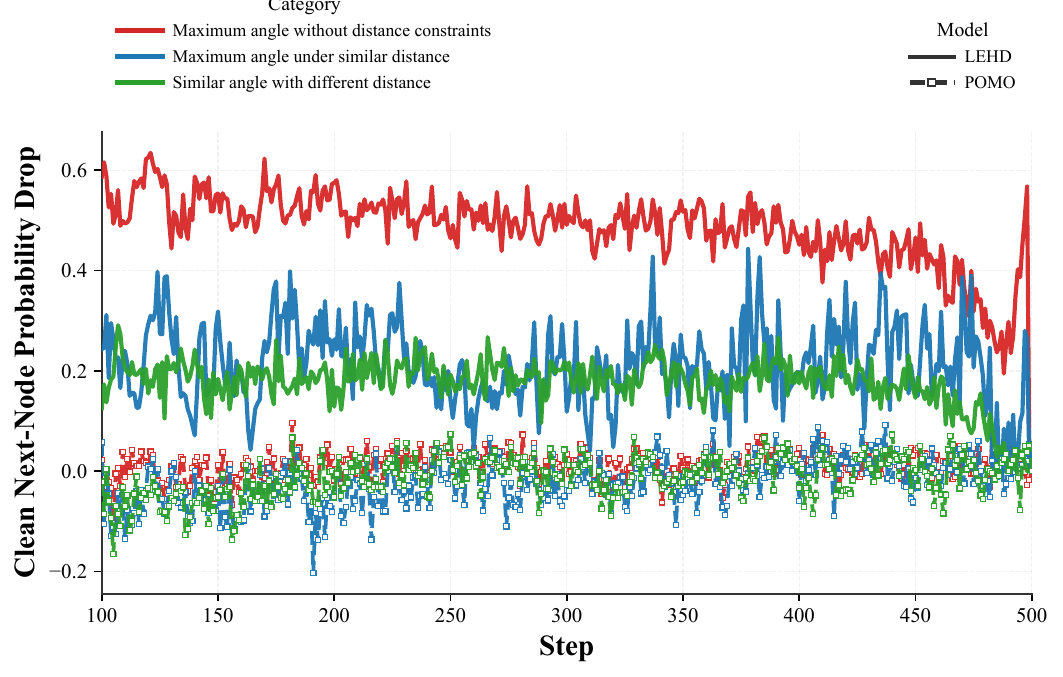}
        \caption{POMO vs. LEHD start-node patching}
        \label{fig:cvrp_start_patching_plot}
    \end{subfigure}
    \caption{
    \textbf{Causal interventions on node representations in CVRP.}
    \textbf{(a)} Probability drop in clean next-node predictions following mean ablation of current versus depot node representations in LEHD.
    \textbf{(b)} Effect of start/depot-node activation patching under three geometric donor conditions, comparing directional sensitivity in LEHD and POMO.
    }
    \label{fig:cvrp_start_node_analysis}
\end{figure}

We next extend our causal intervention studies to CVRP routes. As in the TSP setting, similar causal patterns emerge: \Figref{fig:cvrp_mean_ablation} reveals that LEHD's immediate next-node predictions are far more sensitive to mean-ablating the current node representation than the depot node, while \Figref{fig:cvrp_start_patching_plot} confirms that the angular orientation of the depot node guides global route navigation.

Interestingly, while the Longest Common Subsequence (LCS) trend under depot-node interventions decreases as the donor angle range widens (matching the TSP pattern in Table~1), the absolute LCS values are substantially lower for CVRP (\Tabref{tab:cvrp500-route-order-similarity}). Intervening on the depot node often forces models to switch to entirely different customer clusters, resulting in non-overlapping route memberships and low sequence alignment. Furthermore, because CVRP routes are strictly constrained by vehicle capacity, LEHD cannot rely solely on start-node angular orientation to minimize return costs; it must simultaneously optimize capacity constraints, keeping Reverse LCS values consistently low across all donor angle ranges.

\begin{table}[ht!]
\centering
\caption{\textbf{CVRP-500 route rollout-order similarity under depot-node activation patching.} Results evaluate maximum-angle donor selection without distance constraints. LCS measures forward-order sequence agreement with clean rollouts, while Rev.\ LCS measures agreement with reversed clean rollouts.}
\label{tab:cvrp500-route-order-similarity}
\begin{tabular}{@{}lcc|cc@{}}
\toprule
\multirow{2}{*}{\textbf{Angle Range}} & \multicolumn{2}{c|}{\textbf{LEHD}} & \multicolumn{2}{c}{\textbf{POMO}} \\
\cmidrule(lr){2-3}\cmidrule(lr){4-5}
& \textbf{LCS} & \textbf{Rev.\ LCS} & \textbf{LCS} & \textbf{Rev.\ LCS} \\
\midrule
$0^\circ$--$30^\circ$ & \textbf{0.496} & 0.123 & \textbf{0.168} & 0.079 \\
$30^\circ$--$60^\circ$ & \textbf{0.318} & 0.126 & \textbf{0.108} & 0.065 \\
$60^\circ$--$90^\circ$ & \textbf{0.221} & 0.117 & \textbf{0.080} & 0.064 \\
$90^\circ$--$120^\circ$ & \textbf{0.155} & 0.107 & \textbf{0.080} & 0.071 \\
$120^\circ$--$150^\circ$ & \textbf{0.117} & 0.107 & \textbf{0.104} & 0.083 \\
$150^\circ$--$180^\circ$ & \textbf{0.115} & 0.111 & \textbf{0.151} & 0.118 \\
\bottomrule
\end{tabular}
\end{table}

\subsection{Node-Role Representation Alignment}

A central finding for the TSP was that LEHD projects current and start node representations into canonical, shared regions of the embedding space across decoding steps and instances (Table~2). We evaluate this cross-instance representation alignment for CVRP in \Tabref{tab:cvrp_lehd_role_alignment}. The results confirm that the same canonical representation mechanism operates in CVRP: intermediate and late decoder layers maintain high cosine similarities for the current and depot nodes compared to randomly selected customer nodes.

\begin{table}[ht!]
\centering
\caption{\textbf{Layer-wise cross-instance alignment of node representations in LEHD on Uniform CVRP-100.} Values denote mean cosine similarity ($\pm$ std) across different problem instances and decoding steps for current, depot, and random customer node representations. Enc.\ and Dec.\ denote encoder and decoder layers.}
\label{tab:cvrp_lehd_role_alignment}
\begin{tabular}{lccc}
\toprule
\textbf{Layer} &
\textbf{Current Node} &
\textbf{Depot Node} &
\textbf{Random Node} \\
\midrule
Enc. 1 & $0.513 \pm 0.330$ & $0.514 \pm 0.342$ & $\mathbf{0.514} \pm 0.327$ \\
Dec. 1 & $\mathbf{0.671} \pm 0.251$ & $0.464 \pm 0.296$ & $0.279 \pm 0.383$ \\
Dec. 2 & $\mathbf{0.792} \pm 0.176$ & $0.502 \pm 0.279$ & $0.344 \pm 0.331$ \\
Dec. 3 & $\mathbf{0.772} \pm 0.146$ & $0.684 \pm 0.155$ & $0.395 \pm 0.298$ \\
Dec. 4 & $0.764 \pm 0.135$ & $\mathbf{0.774} \pm 0.113$ & $0.401 \pm 0.307$ \\
Dec. 5 & $0.715 \pm 0.145$ & $\mathbf{0.755} \pm 0.135$ & $0.431 \pm 0.263$ \\
Dec. 6 & $\mathbf{0.735} \pm 0.213$ & $0.721 \pm 0.234$ & $0.732 \pm 0.209$ \\
\bottomrule
\end{tabular}
\end{table}
\clearpage
\section{Experimental Settings}
\label{sec:experiment-settings}

This section summarizes the implementation details and hyperparameter
settings of the experiments.  All models were evaluated using greedy decoding without
model-specific test-time augmentation. All experiments were conducted on a
single NVIDIA GeForce RTX 3090 GPU.

\subsection{Onion-Depth Analysis}
\label{subsec:onion-depth-settings}

For the onion-depth analysis, we set the normalized depth threshold to
$\tau=0.5$. For an onion decomposition discretized into $B$ depth bins,
the minimum depth required for an oscillation is

\begin{equation}
b_{\min}
=
\min\left(B,\left\lfloor 0.5B \right\rfloor+1\right).
\label{eq:onion-min-depth-setting}
\end{equation}

An oscillation is counted when the compressed bin sequence leaves the
outermost bin, reaches at least $b_{\min}$, and subsequently returns to the
outermost bin. Consecutive repetitions of the same bin are removed before
counting.

\subsection{Targeted Start-Node Intervention}
\label{subsec:start-patching-settings}

For the targeted start-node intervention, donor nodes were selected from
the already visited portion of the partial tour. Let $S$ denote the original
start node, $C$ the current node, and $d$ a candidate donor node. We define

\begin{equation}
\theta_d
=
\angle\left(
x_S-x_C,\,
x_d-x_C
\right),
\qquad
r_d
=
\frac{\lVert x_d-x_C\rVert_2}
     {\lVert x_S-x_C\rVert_2}.
\label{eq:donor-geometry}
\end{equation}

The geometric constraints used for the three donor-selection conditions are
reported in Table~\ref{tab:start-donor-settings}.

\begin{table*}[ht!]
\centering
\caption{Geometric constraints used for donor selection in the targeted
start-node intervention.}
\label{tab:start-donor-settings}
\begin{tabular}{@{}lll@{}}
\toprule
Donor condition & Angular constraint & Distance constraint \\
\midrule
Maximum angle without distance constraints
& Maximize $\theta_d$
& None \\

Maximum angle under similar distance
& Maximize $\theta_d$
& $0.80 \leq r_d \leq 1.20$ \\

Similar angle with different distance
& $\theta_d \leq 15^\circ$
& $r_d \leq 0.70$ or $r_d \geq 1.30$ \\
\bottomrule
\end{tabular}
\end{table*}

Thus, similar distance permits a deviation of at most $20\%$ from the
original start-to-current distance. The similar-angle condition restricts
the angular deviation to $15^\circ$, while requiring the donor distance to
be at least $30\%$ smaller or larger.

\subsection{Future-Action Planning Probes}
\label{subsec:future-probe-settings}

For each model layer and prediction horizon, we trained a separate linear
candidate-ranking probe while keeping the underlying routing model frozen.
The data-split and optimization settings are summarized in
Table~\ref{tab:future-probe-training-settings}.

\begin{table}[ht!]
\centering
\caption{Training settings for the future-action planning probes.}
\label{tab:future-probe-training-settings}
\begin{tabular}{@{}lr@{}}
\toprule
Setting & Value \\
\midrule
Training/validation/test split & $70\%/15\%/15\%$ \\
Random seed & 123 \\
Training epochs & 3 \\
Learning rate & $10^{-3}$ \\
Weight decay & $10^{-4}$ \\
Gradient clipping norm & $1.0$ \\
Fallback batch size & 16 \\
\bottomrule
\end{tabular}
\end{table}

\subsection{Training Settings}\label{subsec:training-settings}

For the standard benchmark experiments, we used the official checkpoints released by the authors to ensure that our baseline results faithfully match the original implementations. To analyze the effect of the training data distribution, we retrained each model on clustered instances while keeping its architecture and original training paradigm unchanged.

We did not switch any model to a different training paradigm (e.g., RL to SL or SL to RL). Such a switch is not straightforward and would change the method under comparison. AM and POMO are originally RL methods: AM is trained with REINFORCE using a rollout baseline, while POMO uses REINFORCE with multiple starting nodes and a shared baseline. These mechanisms are specific to the RL formulation, and adapting them to SL would require a non-trivial redesign of the training objective and targets. For LEHD, the original method is SL; the authors explicitly note that RL training is impractical because the heavy decoder structure incurs substantial memory and computational costs. We therefore restrict our experiments to each model's original training setting, which allows us to isolate the effect of the training distribution rather than confounding it with a change in the learning algorithm.
\clearpage

\end{document}